\documentclass[lettersize,journal]{IEEEtran}

\ifCLASSINFOpdf
\else
   \usepackage[dvips]{graphicx}
\fi
\usepackage{url}

\usepackage{graphicx} 
\usepackage[top=0.756in, bottom=1.033in, right=0.635in, left=0.635in, paper=letterpaper]{geometry}
\usepackage[linesnumbered,ruled,vlined]{algorithm2e}
\SetKwInput{KwInput}{Input}
\SetKwInput{KwOutput}{Output}

\SetKwFor{For}{for}{do}{}
\SetKwFor{parfor}{for}{do in parallel}{}

\usepackage{multirow}
\usepackage{subcaption}
\usepackage{amssymb}
\usepackage{amsmath}
\usepackage{hyperref}
\usepackage{dsfont}
\usepackage{xcolor}
\usepackage{setspace}
\usepackage{float}
\usepackage{enumitem}

\usepackage{array}
\usepackage{caption}
\usepackage{booktabs}

\usepackage{comment}
\usepackage{amsmath}  
\usepackage{graphicx} 
\usepackage{amsthm}
\usepackage{balance}

\theoremstyle{definition}
\newtheorem{theorem}{Theorem}

\renewcommand{\thesection}{\Roman{section}}

\allowdisplaybreaks

\makeatletter
\def\BState{\State\hskip-\ALG@thistlm}
\makeatother
\usepackage[font=footnotesize,labelfont=bf]{caption}
\begin{document}

\title{Cooperative Decentralized Backdoor Attacks \\ on Vertical Federated Learning
}


\author{Seohyun Lee,~\IEEEmembership{Student Member,~IEEE}, ~Wenzhi Fang,~\IEEEmembership{Student Member,~IEEE}, \\ Anindya Bijoy Das,~\IEEEmembership{Member,~IEEE}, ~Seyyedali Hosseinalipour,~\IEEEmembership{Senior Member,~IEEE}, \\
~David J. Love,~\IEEEmembership{Fellow,~IEEE}, ~Christopher G. Brinton,~\IEEEmembership{Senior Member,~IEEE}

\thanks{S. Lee, W. Fang, D. Love, and C. Brinton are with the Elmore Family School of Electrical and Computer Engineering, Purdue University, West Lafayette, IN, 47907. Email: \{lee3296, fang375, djlove, cgb\}@purdue.edu.}
\thanks{A. Das is with the Department of Electrical and Computer Engineering, University of Akron, Akron, OH, 44325. Email: adas@uakron.edu.}
\thanks{S. Hosseinalipour is with the Department of Electrical Engineering, University at Buffalo-SUNY, Buffalo, NY, 14260. Email: alipour@buffalo.edu.}
\thanks{This work was supported in part by the Office of Naval Research (ONR) under grants N00014-22-1-2305 and N00014-21-1-2472, and by the National Science Foundation (NSF) under grant CNS-2212565.}
}


\IEEEtitleabstractindextext{%

\begin{abstract}
Federated learning (FL) is vulnerable to backdoor attacks, where adversaries alter model behavior on target classification labels by embedding triggers into data samples. While these attacks have received considerable attention in horizontal FL, they are less understood for vertical FL (VFL), where devices hold different features of the samples, and only the server holds the labels. In this work, we propose a novel backdoor attack on VFL which (i) does not rely on gradient information from the server and (ii) considers potential collusion among multiple adversaries for sample selection and trigger embedding. Our label inference model augments variational autoencoders with metric learning, which adversaries can train locally. A consensus process over the adversary graph topology determines which datapoints to poison. We further propose methods for trigger splitting across the adversaries, with an intensity-based implantation scheme skewing the server towards the trigger. Our convergence analysis reveals the impact of backdoor perturbations on VFL indicated by a stationarity gap for the trained model, which we verify empirically as well. We conduct experiments comparing our attack with recent backdoor VFL approaches, finding that ours obtains significantly higher success rates for the same main task performance despite not using server information. Additionally, our results verify the impact of collusion on attack performance.
\end{abstract}
\vspace{-2mm}
}

\maketitle
\IEEEdisplaynontitleabstractindextext
\IEEEpeerreviewmaketitle

\section{Introduction}
\label{sec:intro}
Federated Learning (FL) \cite{mcmahan2017communication} has emerged as a popular method for collaboratively training machine learning models across edge devices. 
By eliminating the need for communication of raw data across the network, FL proves especially valuable in scenarios where data privacy is critical.
However, the decentralized nature of FL introduces new security challenges, as individual devices may lack the robust security measures of a centralized system, thereby increasing the risk of adversarial attacks that can compromise the integrity of training.

The two prevalent frameworks of FL, Horizontal Federated Learning (HFL) and Vertical Federated Learning (VFL), both face significant vulnerabilities to adversaries \cite{bouacida2021vulnerabilities}.
In HFL, the training data is partitioned by sample data points, with each device holding different subsets of the overall dataset. 
Most of the existing literature on adversarial attacks in FL has concentrated on HFL, with the goal to tackle vulnerabilities such as data poisoning attacks \cite{bhagoji2019analyzing, ma2019data}, model inversion attacks \cite{masuda2021model, issa2024rve}, and backdoor attacks \cite{bagdasaryan2020backdoor}. 
Conversely, VFL \cite{wei2022vertical, romanini2021pyvertical} involves local devices that share the same samples but hold different features of the samples.
In this setup, one node, referred to as the active party or server, holds the labels and oversees the aggregation process, while the other devices function as passive parties or clients, constructing local feature embeddings and periodically passing them to the server.
For instance, in a wireless sensor network (WSN) \cite{mengistu2024survey}, each sensor may collect readings from its local environment (e.g., video feeds) which collectively form a full sample for the fusion center's learning task (e.g., object detection) at a point in time. 

Recent research has begun to study the impact of attacks on VFL, including feature inference attacks \cite{yang2023practical,luo2021feature}, label inference attacks \cite{fu2022label}, and attribute inference attacks \cite{Song2019OverlearningRS}.
In this paper, we focus on \textit{backdoor attacks} \cite{bagdasaryan2020backdoor} for VFL. A backdoor attack aims to alter the behavior of an FL model on a particular label (called the \textit{target}) when the model encounters data samples for the label that an adversary has implanted with an imperceptible trigger pattern.
Addressing backdoor attacks is crucial in both HFL and VFL because these attacks can lead to severe security breaches without easily detectable impacts on overall model performance \cite{lyu2023poisoning}. For example, in the WSN object detection use-case, a backdoor adversary could implant triggers in a sensor's local sample view of a car to force the system to misclassify the entire sample as a truck. 

In this paper, we consider an underexplored scenario where there are multiple adversaries, and these adversaries have the capability of colluding over a graph topology to execute a coordinated backdoor attack. In this setting, the adversaries gain control of a set of client nodes and establish an ad-hoc local network among themselves for cooperation. Prominent examples can be found in defense settings. For instance, in a contested region, attackers could take control of a few scattered military assets responsible for e.g., communicating front-line conditions, such as drones or tactical mobile devices \cite{luo2018attribute}. Such control could be achieved through a variety of means, for example, exploiting vulnerabilities over control links to hijack drones \cite{pratama2024behind} or injecting malicious malware onto the devices \cite{gulatas2023malware}. Taking advantage of this, the adversaries could begin colluding to analyze and transmit data across a local adversary-formed network topology. To form this graph, the adversaries could employ device-to-device (D2D) communication protocols, which have proliferated in 5G wireless and have shown benefit in distributed learning, allowing devices to synchronize model parameters and/or gain a better estimate of the overall data distribution \cite{wagle2025multi, lee2024smart, xing2020decentralized}. This decentralized topology allows for the exchange of information between adversaries that might not be readily available at an individual level, i.e., a more complete view of each sample. By pooling their individual observations, they can potentially infer sensitive information about the battlefield--such as troop numbers, logistics capabilities, or defensive positions. Moreover, this making it easier for them to mislead the central decision-making system via a backdoor attack (e.g., inducing the command center to misclassify an enemy fighter as a benign node).

Still, the adversaries might be unable to engage in maximal information exchange (i.e., forming a full mesh graph) due to e.g., resource constraints, geographical distances, and channel conditions preventing formation of certain D2D links. We thus need to understand how adversarial collusion impacts the attack potency, and the role played by adversarial connectivity. Moreover, following the common assumption in the literature that the adversaries have full control over the compromised node(s) \cite{lin2019byzantine}, it is important to consider the full range of adversarial capabilities. By doing so, we do not underestimate the adversary’s capabilities and insights, which could otherwise lead to overconfidence in the system’s security.

\subsection{Related Work}
\label{ssec:related}
Extensive research has been conducted on backdoor attacks in HFL. In these cases, adversaries send malicious updates to the server, causing the model to misclassify data when a trigger is present without impacting the overall performance of the FL task \cite{fung2018mitigating, wang2020attack, cao2022highly}. In this domain,
\cite{bagdasaryan2020backdoor} proposed a scale-and-constrain methodology, in which the adversary's local objective function is modified to maximize attack potency without causing degradation of the overall FL task. 
\cite{xie2019dba} explored trigger embeddings that take advantage of the distributed nature of HFL, by dividing the trigger into multiple pieces. 
In addition to the various attacks, defenses for these vulnerabilities in HFL have also been studied, e.g., \cite{nguyen2022flame, Fung2020}. Another significant issue with the effectiveness of backdoor attacks in HFL is the presence of non-i.i.d. data distributions across local clients, resulting in slower convergence of the global model \cite{hsieh2020non}. In this regard, an HFL backdoor methodology \cite{han2024badsfl} has been developed to work with a popular HFL algorithm called SCAFFOLD \cite{karimireddy2020scaffold} by utilizing Generative Adversarial Networks (GANs) \cite{goodfellow2020generative}.

In our work, we focus on backdoor attacks for VFL, which have not been as extensively studied.
The VFL scenario introduces unique challenges: devices do not have access to sample labels or a local loss function, and must rely on gradients received from the server to update their feature embedding models. Thus, unlike in the HFL scenario, where attackers can utilize a ``dirty-label’’ backdoor by altering the labels on local datapoints \cite{gu2019badnets}, an attack in VFL must be a ``clean-label’’ backdoor attack, since only the server holds the training labels. 

In this context, a few recent techniques have been developed to carry out backdoor attacks in VFL through different methods for inferring training sample labels. These include using gradient similarity \cite{xuan2023practical} and gradient magnitude \cite{bai2023villain} comparisons with a small number of reserve datapoints the adversary has labels for.  In this domain, \cite{fu2022label} created a local adaptive optimizer that changes signs of gradients inferred to be the target label. In a similar vein, other works have exploited the fact that the indices of the target label in the cross-entropy loss gradients will have a different sign, provided that the model dimension matches the number of classes \cite{liu2020backdoor, 10296882}. Other works have also considered gradient substitution alignment to conduct the backdoor task with limited knowledge about the target label \cite{chen2023practical}. Moreover, researchers have explored the implications of backdoor attacks in different settings, such as with graph neural networks (GNNs) \cite{yang2024backdoor}. Further, \cite{naseri2023badvfl} considered training an auxiliary classifier to infer sample labels based on server gradients.

Despite these recent efforts, a major limitation of the existing approaches is that \textit{they rely on information sent from the server to conduct the backdoor attack, in addition to using it for VFL participation}. 
This dependency can enable the server to implement defense mechanisms, particularly during the label inference phase, which can significantly limit the effectiveness of the attack \cite{liu2024vertical}. While research  on bypassing the use of server-received information in backdooring VFL exists, it is limited to only \textit{binary classification tasks \cite{chen2023universal}}.
Additionally, the aforementioned studies \cite{xuan2023practical, bai2023villain, liu2020backdoor, 10296882, chen2023universal} \textit{concentrate on the classic two-party VFL scenario with a single adversary}, which fits them into the cross-silo FL context \cite{romanini2021pyvertical}, leading to a limited understanding of VFL backdoors in networks where multiple adversaries may collude to carry out the attack. 
In particular, unlike cross-silo FL, which typically involves a few participants such as large organizations, cross-device FL often encompasses numerous distributed devices collaborating to construct a global model, thereby significantly increasing the potential for security malfunctions \cite{mcmahan2017communication, bagdasaryan2020backdoor}.

\vspace{-0.05in}
\subsection{Research Questions and Approach Overview}
These limitations lead us to investigate the following two research questions (RQs) in this work:
\begin{itemize}
    \item \textbf{RQ1: }Can an adversary successfully implement a backdoor injection into the server's VFL model using only locally available information for label inference?
    \item \textbf{RQ2: }How can multiple adversaries collude with limited sharing to construct a backdoor injection in cross-device VFL, and what is the impact of their graph connectivity?
\end{itemize}

\noindent \textbf{Overview of approach.} We develop a novel backdoor VFL strategy that addresses the above questions. To answer RQ1, we introduce a methodology for an adversary to locally infer and generate datapoints of the target label for attack. Our approach leverages Variational Autoencoders (VAE) \cite{kingma2013auto} and triplet loss metric learning \cite{schroff2015facenet} to determine which samples should receive trigger embeddings, avoiding leveraging server gradients.
To answer RQ2, we employ the graph topology of adversarial devices to conduct cooperative consensus on which samples should be implanted with triggers. In this regard, we show both empirically (Fig. \ref{fig:graph-conn} \& Fig. \ref{fig:deltafunc}) and theoretically (Sec. \ref{theorem}) that the effectiveness of the attack and gradient perturbation is dependent on this graph topology. We also develop an intensity-based triggering scheme and two different methods for partitioning these triggers among adversaries, leading to a more powerful backdoor injection than existing attacks.

\vspace{-0.21cm}
\subsection{Outline and Summary of Contributions} 
\begin{itemize}[leftmargin=4mm]
    \item We propose a novel collaborative backdoor attack on VFL which does not rely on information from the server.
    Our attack employs a VAE loss structure augmented with metric learning for each adversary to independently acquire its necessary information for label inference. 
    Following the local label inference, the adversaries conduct majority consensus over their graph topology to agree on which datapoints should be poisoned (Sec. \ref{sec:inference}\&\ref{sec:method-label}). 

    \item For trigger embedding, we develop an intensity-based implantation scheme which brings samples closer to the target without compromising non-target tasks. Attackers employ their trained VAEs to generate new datapoints to be poisoned that are similar to the target label, forcing the server to rely more on the embedded trigger. Adversary collusion is facilitated via two proposed methods for trigger splitting, either subdividing one large trigger or embedding multiple smaller triggers (Sec.~\ref{sec:backdoor}\&\ref{sec:trigger}).

    \item We conduct convergence analysis of cross-device VFL under backdoor attacks, revealing the degradation of main task performance caused by adversaries. Specifically, we show that the server model will have a stationarity gap proportional to the level of adversarial gradient perturbation (Sec. \ref{theorem}). We provide an interpretation for this gradient perturbation as an increasing function of the adversary graph's algebraic connectivity and average degree, which we further investigate empirically (Sec. \ref{sec:simulations}). We are unaware of prior works with such convergence analysis on VFL under backdoor attacks.
    

    \item We conduct extensive experiments comparing the performance of our attack against the state-of-the-art \cite{xuan2023practical, bai2023villain} on five image classification datasets. Our results show that despite not using server information, we obtain a $30\%$ higher attack success rate for comparable main task performance. We also show an added advantage of our decentralized attack in terms of  improved robustness to noising defenses at the server. We also demonstrate that higher adversarial graph connectivity yields improved attack success rate with our method, thus corroborating our theoretical claims (Sec.~\ref{sec:simulations}).
\end{itemize}
\newtheorem{assumption}{\textbf{Assumption}}

\section{System Model}

\begin{table}[t]
    \vspace{2mm}
    \begin{center}
        \footnotesize
        \setlength{\tabcolsep}{4pt} 
        \caption{\small Summary of main notations employed throughout the paper.\vspace{-.5mm}}
        \label{table:variables}
        \begin{tabular}{|c|p{0.39\textwidth}|} 
        \hline
        \textbf{Notation} & \multicolumn{1}{|c|}{\textbf{Description}} \\ 
        \hline
        $k$ & Any client, excluding the server $K$ \\ \hline
        $m$ & An adversary client \\ \hline
        $\mathcal{K}$ & The set of all clients, including the server \\ \hline
        $\mathcal{A}$ & The set of all adversary clients \\ \hline
        $G$ & The graph formed amongst adversary clients $m \in \mathcal{A}$ \\ \hline
        $\mathcal{A}_m$ & A subset of $\mathcal{A}$, the neighbors to adversary $m$ in graph $G$ \\ \hline
        \raisebox{-1.2ex}[0pt]{$x_{k}^{(i)}$} & The feature partition belonging to local client $k$ for sample datapoint $x^{(i)}$ \\ \hline
        \raisebox{-1.2ex}[0pt]{$X_m^{(i)}$} & The concatenated samples for adversary $m$ received over edges in $G$ \\ \hline
        $\widetilde{x}_{m}^{(i)}$ & Sample generated from adversary $m$'s VAE \\ \hline
        \raisebox{-1.2ex}[0pt]{$\widehat{x}_{m}^{(i)}$} & $\widetilde{x}_{m}^{(i)}$ implanted with the trigger pattern subportion corresponding to adversary $m$ \\ \hline
        $z_m^{(i)}$ & Latent variable produced from VAE encoder \\ \hline
        $\mathcal{D}$ & The overall dataset without being partitioned amongst clients \\ \hline
        $\mathcal{D}_m$ & A subset of $\mathcal{D}$ containing concatenated datapoints $X_m^{(i)}$ \\ \hline
        \raisebox{-1.2ex}[0pt]{$\widehat{\mathcal{D}}_m$} & A subset of $\mathcal{D}_m$, concatenated samples of only those where the label is known \\ \hline
        \raisebox{-1.2ex}[0pt]{$\widehat{\mathcal{D}}_m^{\text{target}}$} & A subset of $\widehat{\mathcal{D}}_m$, concatenated datapoints belonging only to target label out of known datapoints \\ \hline
        $\mathcal{D}_{m}^{(p)}$ & Locally inferred datapoints for adversary $m$ \\ \hline
        $\mathcal{D}_{g}^{(p)}$ & Collaborative inference set based off $\{\mathcal{D}_{m}^{(p)} | m \in \mathcal{A} \}$ \\ \hline
        $\widetilde{\mathcal{D}}_{m}^{(p)}$ & $m$'s feature partition slice of $\mathcal{D}_{g}^{(p)}$ \\ \hline
        \raisebox{-1.1ex}[0pt]{$f_k$} & The local model of client $k$, producing feature embeddings to be sent to server $K$ \\ \hline
        $\phi_{K}$ & The server model for server $K$ \\ \hline
        \raisebox{-1.2ex}[0pt]{$\theta_{k}$} & The parameters for local models $f_{k}$. For server $K$, its server model is parameterized as $\theta_{K}$ \\ \hline
        $\phi_{m}^{\text{VAE}}$ & Adversary $m$ local VAE \\ \hline
        $\mu$ & The mean vector of an adversarial VAE \\ \hline
        $\phi_{m, \mu}$ & Auxiliary classifier for adversary $m$ \\ \hline
        $\zeta$ & The poisoning budget \\ \hline
        $y_t$ & The target label \\ \hline
        $\rho$ & The connectivity of the graph $G$ \\ \hline
        $\delta(\rho)$ & The gradient perturbation from adversaries for connectivity $\rho$ \\ \hline
        $\widehat{\kappa}$ & The margin value of the triplet loss \\ \hline
        $a$ & Anchor datapoint for the triplet loss \\ \hline
        $p$ & Positive datapoint for the triplet loss \\ \hline
        $n$ & Negative datapoint for the triplet loss \\ \hline
        \end{tabular}
    \end{center}
    \vspace{-3mm}
\end{table}

\subsection{Vertical Federated Learning Setup}
We consider a network of $K$ nodes within the vertical federated learning (VFL) setup collected in the set $\mathcal{K} = \{1, 2, ..., K\}$, where $k = 1, \dots, K - 1$ are the clients and $k = K$ is the server. 
We assume a black-box VFL scenario, where the clients do not have any direct knowledge about the server and global objective, e.g., the model architecture, loss function, etc.
We denote the overall dataset as $\mathcal{D}$, and the total number of datapoints is $N = |\mathcal{D}|$. Each client contains a separate disjoint subset of datapoint features. We represent the $i^{th}$ datapoint of $\mathcal{D}$ as $x^{(i)} = \{x_{1}^{(i)},\dots,x_{_{K-1}}^{(i)}\}$, where $x_{k}^{(i)}$ belongs to the local data of client $k = 1, 2, \dots, K - 1$. Note that only the server $K$ holds the labels $\mathcal{Y} = \{y_1,\dots, y_{_N}\}$ associated with the corresponding dataset. 

Each client locally trains its feature encoder on its data partition, and the server is responsible for coordinating the aggregation process.
In particular, we adopt a Split Neural Network-based (SplitNN) VFL setup \cite{romanini2021pyvertical}, where the clients send their locally produced feature embeddings (sometimes called the bottom model) to the server. 
The server then updates its global model (the top model) and returns the gradients of the loss with respect to the feature embeddings back to the clients. 

Mathematically, the optimization objective of the VFL system can be expressed as
\vspace{-0.2cm}
\begin{align}
\begin{split}
    \min_{{\theta}_{1}, \ldots, {\theta}_{_K}} & F({\theta}) :=  \frac{1}{N} \sum_{i=1}^{N} \mathcal{L}  \left(y_i, \phi_{_K}\left( \{f_1(x_1^{(i)};\theta_1), \right. \right. \\
     & \left. \left. f_2(x_2^{(i)};\theta_2), \ldots, f_{K\!-\!1}(x_{K\!-\!1}^{(i)};\theta_{K\!-\!1})\};{\theta}_{_K} \right) \right), \label{eq:1}
\end{split}
\end{align}
where $f_{k}$ denotes the embedding mapping function of client $k \in \mathcal{K}\setminus \{K\}$ parameterized by $\theta_k$, $\phi_{_K}$ is the server model parameterized by $\theta_{_K}$, ${\theta} = \{ \theta_1, \ldots, \theta_K \}$, and $\mathcal{L}$ denotes the loss function of the learning task. 

\begin{figure}[t] 
    \centering
        \includegraphics[width=\linewidth]{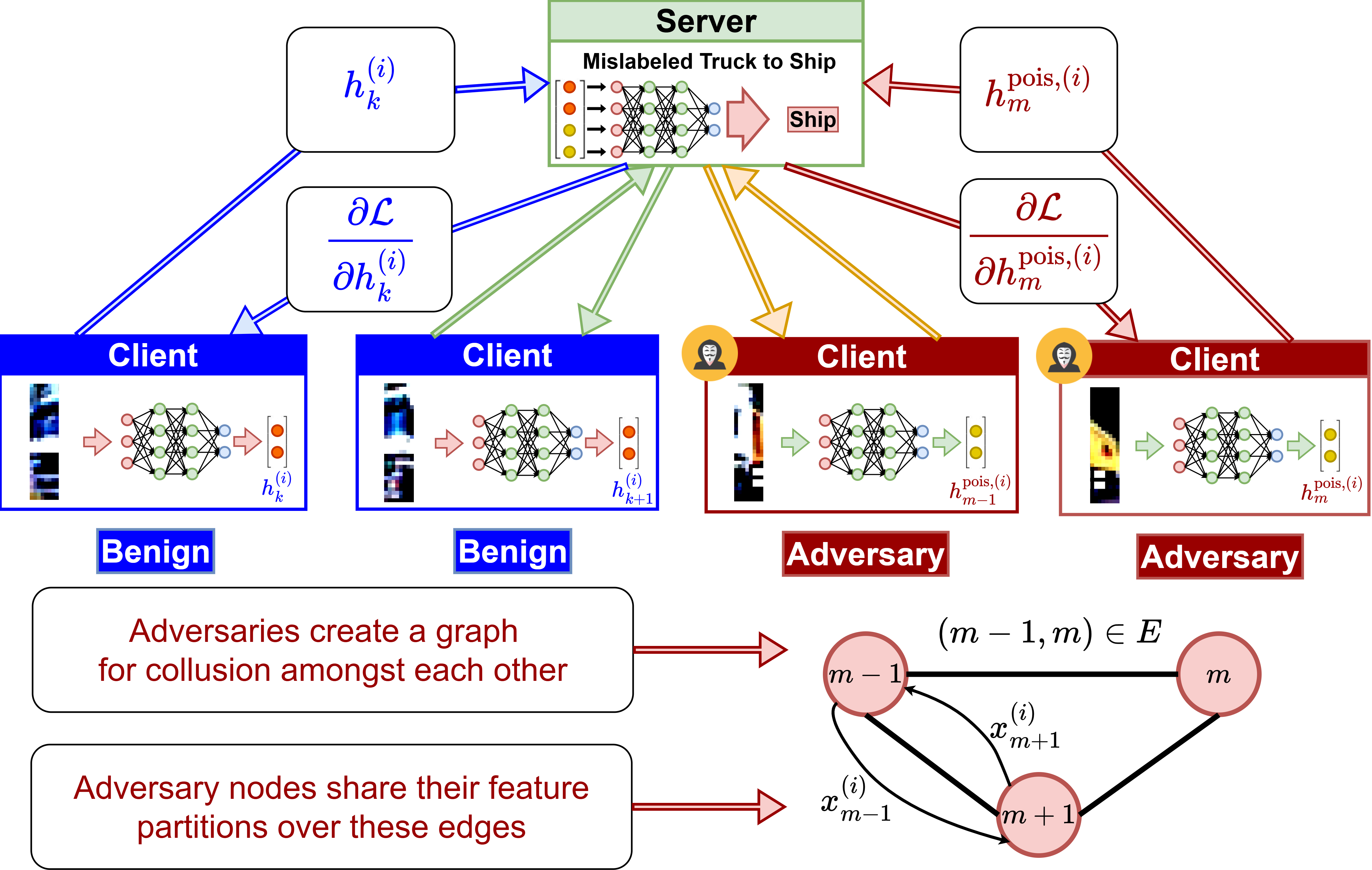}
        \caption{Client-server sharing of embeddings and gradients in VFL. An example of a feature-partitioned datapoint undergoing a backdoor trigger implantation is shown. The adversaries send up their poisoned embeddings, which are then concatenated by the server and cause misclassification. Moreover, adversaries form a graph amongst each other, sharing their feature partitions to enhance insights on the samples they wish to poison.}
        \label{fig:vfl-poison-embed}
        \vspace{-0.2in} 
\end{figure}

To solve~\eqref{eq:1}, in each VFL training round $t$, the server selects a set of mini-batch indices. Across clients, the full mini-batch set is $\mathcal{B}^{(t)} = \{(x_{1}^{(i)},...,x_{K-1}^{(i)})| x_{k}^{(i)} \in \mathcal{B}^{(t)}_k \} \subset \mathcal{D}$. Each client $k$ needs to update its own local model $\theta_k$ on its mini-batch subset $\mathcal{B}_k^{(t)}$. 
However, different from HFL, the gradients of the clients' loss function models depend on information from the server, while the model update at the server also depends on the mapping computed by the clients. 
Thus, during round $t$, each client $k$ first computes local low-dimensional latent feature embeddings \(H_k^{(t)} = \{h_k^{(i)} = f_k(x_k^{(i)};\theta_k^{(t)}) | x_{k}^{(i)} \in  \mathcal{B}_k^{(t)}\}\). 
After the embeddings $H_k^{(t)}$ are obtained, as seen in Fig. \ref{fig:vfl-poison-embed}, they are sent to the server. The server computes the gradient $\frac{\partial \mathcal{L}}{\partial \theta_{_K}}$ to update the top model $\theta_{_K}$ via gradient descent: $\theta_{_K}^{(t+1)} \leftarrow \theta_{_K}^{(t)} - \eta_{_K}^{(t)} \frac{\partial \mathcal{L}}{\partial \theta_{_K}}$.
In addition, the server computes $\frac{\partial \mathcal{L}}{\partial h_k^{(i)}}$, 
which is sent back to client $k$ for the computation of gradients of the loss function with respect to the local model $\theta_k^{(t)}$ as 
$\frac{\partial \mathcal{L}}{\partial \theta_k}  = \frac{1}{|\mathcal{B}_k^{(t)}|} \sum_{h_k^{(i)} \in H_k^{(t)}} \frac{\partial \mathcal{L}}{\partial h_k^{(i)} } \frac{\partial h_k^{(i)}}{\partial \theta_k}$.
The client then updates its model via gradient descent: $\theta_{k}^{(t+1)} \leftarrow \theta_{k}^{(t)} - \eta_{k}^{(t)} \frac{\partial \mathcal{L}}{\partial \theta_{k}}$.

\subsection{Backdoor Attacks in VFL}
\label{sec:intro-backdoor}

In this work, we investigate backdoor attacks on VFL, where each adversary is a client in the system (i.e., compromised client). 
The goal of the adversaries is to modify the server model's behavior on data samples of a target label (i.e., the label that adversaries want to induce misclassification on) via an implanted backdoor trigger on these samples.
Importantly, however, the model should still perform well on clean data for which the trigger is not present. 
While modifying the objective function is a common method for backdoor attacks in the HFL case \cite{bagdasaryan2020backdoor}, this is not feasible in VFL because only the server can define the loss, and hence the adversaries must follow the server's loss function. 
Therefore, we consider attacks where an adversary $m$ implants a trigger $\sigma_{m}$ into a selected datapoint $i$ \textit{inferred} to be of the target label, i.e., producing $x_m^{(i)} + \sigma_m$.

In addition, we assume that multiple adversarial clients $\mathcal{A} \subset \mathcal{K}$ can collude to plan the attack. In this vein, we consider a connected, undirected graph $G = (\mathcal{A}, E)$ among the adversaries, where $E$ denotes the set of edges. For adversary $m \in \mathcal{A}$, we denote $\mathcal{A}_m = \{m' : (m, m') \in E\}$ as its set of neighbors.
Adversaries will employ $G$ to conduct collaborative label inference, as will be described in Sec. \ref{sec:inference}\&\ref{sec:method-label}. 

\section{Attack Methodology}
\label{sec:methodology}
To execute the backdoor attack in VFL, adversaries need to (i) identify datapoints belonging to the target label (Sec.~\ref{sec:inference}) and (ii) implant triggers on the corresponding datapoints to induce misclassification (Sec.~\ref{sec:backdoor}). We present the methodology for these processes in this section, and give more specific algorithmic procedures in Sec.~\ref{sec:algorithm}.

\subsection{Label Inference}
\label{sec:inference}
Our label inference methodology is summarized in Fig.~\ref{fig:system}:
\begin{figure*}[t!] 
    \centering
        \includegraphics[width=\linewidth]{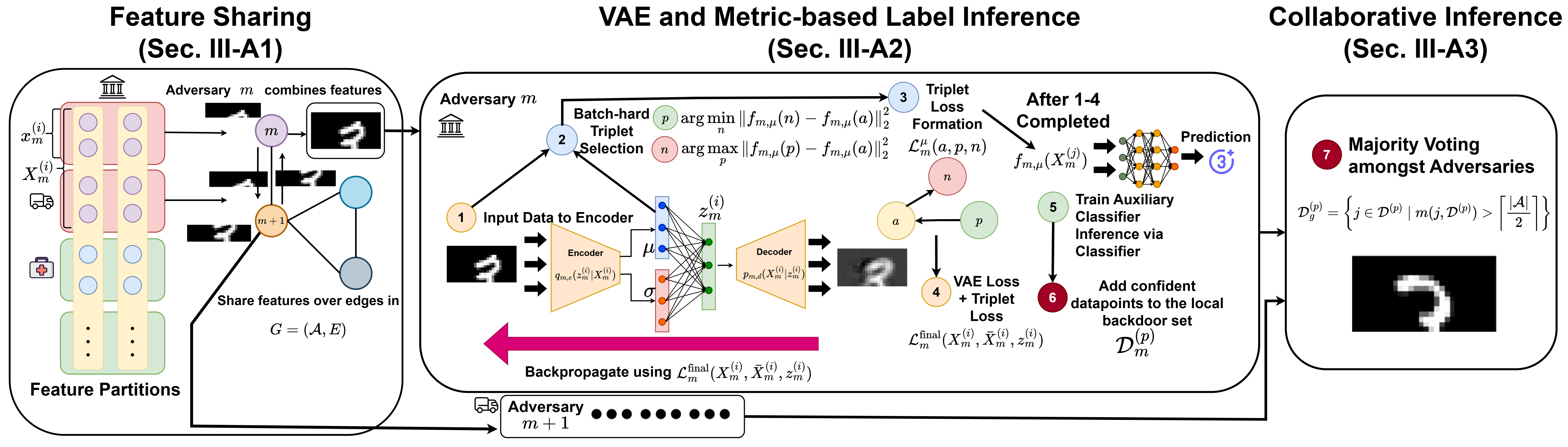}
        \vspace{-5.65mm}
        \caption{Label inference methodology with modified VAE architecture. Initially, the adversaries share their feature partitions. At the $\mu$ layer of the VAE, triplet margin loss is employed to conduct metric learning via the known label datapoints. After training the VAE, the $\mu$ vectors are used to perform a classification task for inference of the target label, with the results of local inference being used in a majority voting scheme for a final collaborative inference of the indices.}
    \label{fig:system}
        \vspace{-6mm}
\end{figure*}


\subsubsection{\textbf{Feature Sharing}}
\label{subsubsec:feature-share}
As in existing work \cite{bai2023villain,fu2022label}, we assume that the adversaries possess labels for a small (e.g., $<1$\%) set of the datapoints.
Even so, when numerous clients each hold a small feature partition of the samples (Feature partition block of Fig. \ref{fig:system}), extracting meaningful information without employing gradients from the server (which we aim to avoid, as discussed in Sec.~\ref{sec:intro}) becomes challenging. This difficulty arises due to the presence of irrelevant features within sample partitions, e.g., a blank background.

To address this issue, the adversaries utilize their collusion graph $G$ discussed in Sec. \ref{sec:intro-backdoor}, through which they exchange feature partitions of their datapoints with their one-hop neighbors (Feature Sharing phase of Fig. \ref{fig:system}). Each adversary $m$ concatenates the partitions as $X_m^{(i)} = \bigcup_{m' \in \mathcal{A}_m \cup m} x_{m'}^{(i)}$. We denote this dataset as $\mathcal{D}_m \subset \mathcal{D}$, a further subset $\widehat{\mathcal{D}}_m \subset \mathcal{D}_m$ of which is for known labels.
\subsubsection{\textbf{VAE and Metric-based Label Inference}}
Next, using $\mathcal{D}_m$, each adversary will conduct local label inference (VAE and Metric-based Label Inference phase of Fig. \ref{fig:system}). We propose leveraging Variational Autoencoders (VAE) as a framework for this, deploying one VAE model $\phi_{m}^{\text{VAE}}$ on each adversary device.
Unlike their AE counterparts \cite{zhai2018autoencoder}, VAEs are simpler to use for generative purposes, as a variable $z$ sampled from the VAE's latent space can be fed through the decoder to generate new datapoints \cite{kingma2013auto}.
To do this, we assume that each datapoint $X_m^{(i)}$ is generated from latent variables $z_m^{(i)}$ following a distribution $p(z_m^{(i)})$, which usually is a standard normal distribution, $\mathcal{N}(0, I)$.
Therefore, the goal of the VAE's decoder model is to learn its parameters to maximize {\small $p_{m, d}(X_m^{(i)}|z_m^{(i)})$}.

However, {\small$p(X_m^{(i)}) = \int p_{m, d}(X_m^{(i)}|z_m^{(i)})p(z_m^{(i)})dz_m^{(i)}$} is computationally intractable, making it unrealistic to calculate the term directly.
Therefore, rather than maximizing $p_{m, d}$ directly, the VAE employs its encoder $q_{m, e}$ as an approximate model which outputs a mean $\mu$ and standard deviation $\sigma$,
reducing the latent space to a univariate Gaussian $\mathcal{N}(\mu, \sigma^2)$. 
The error can be captured in a KL divergence-based objective \cite{kullback1951information} which measures the difference between two probability distributions, $q_{m, e}(z_m^{(i)})$ and $ p(z_m^{(i)})$, denoted as 
$D_{\text{KL}}(q_{m, e}(z_m^{(i)}) \| p(z_m^{(i)}))$.
This term, when included in the loss function, can encourage the latent space to be closer to a standard normal distribution, allowing for random sampling from the latent space for datapoint generation.

Additionally, the VAE aims to optimize its reconstruction loss. We adopt the mean-squared error (MSE) metric 
$\mathcal{L}_m^{\text{rec}}(x, \bar{x}_m(x)) = \| {x} - \bar{x}_m(x) \|_2^2$,
where $\bar{x}_m(x)$ is the output reconstructed by the decoder for input $x$. Combining these together, a typical VAE trains for any datapoint $x$ on the objective function
\begin{equation}
\begin{split}
    \mathcal{L}_m^{\text{VAE}}(x, \bar{x}_m(x), z) =& \lambda \cdot \mathcal{L}_m^{\text{rec}}(x, \bar{x}_m(x)) \\ 
     +& (1-\lambda) \cdot D_{\text{KL}}(q_{m, e}(z) \| p(z)), \label{eq:5}
\end{split}
\end{equation}

\noindent where $0 < \lambda < 1$ captures the importance weight of each individual term.

A key advantage of a VAE is its ability to utilize the latent space to learn separable embeddings.
Additionally, existing work has hypothesized that applying metric learning to the $\mu$ vector can enhance embedding alignment within the latent space \cite{ishfaq2018tvae}.
We leverage this by training a joint triplet margin loss \cite{schroff2015facenet} objective alongside the standard VAE, given by
 \begin{equation}
    \begin{split}
        \mathcal{L}_{m}^{\mu}(a,p,n) = \max \left( d_{m, \mu}^2(a, p) - d_{m, \mu}^2(a, n) + \widehat{\kappa}, 0 \right), \label{eq:6}
    \end{split}
\end{equation}

\noindent where $d_{m, \mu}(a,r) = \left \Vert f_{m, \mu}(a)-f_{m, \mu}(r) \right \Vert_{2}$. Here, $a$ is the anchor datapoint, $p$ is a datapoint with the same label (called the \textit{positive}), $n$ is a datapoint belonging to a different label (called the \textit{negative}), $\widehat{\kappa}$ is the margin hyperparameter, and $f_{m, \mu}(\cdot)$ is the function induced by the $\mu$ vector of the VAE.
The triplet margin loss creates embeddings that reduce the distance between the anchor $a$ and the positive $p$ in the feature space while ensuring the negative $n$ is at least a distance $\widehat{\kappa}$ from $p$. 

Now, using the labeled dataset $\widehat{\mathcal{D}}_m$, the positives and anchors are the set of datapoints belonging to the target label, and negatives are from the other labels. However, as outlined in \cite{schroff2015facenet}, careful triplet selection is required for a good embedding alignment. Therefore, we employ the ``batch-hard" method of online triplet selection \cite{li2021unified}, where the ``hardest'' positive and negative are chosen. These include the farthest positive and the closest negative to the anchor embedding, given by $\widehat{p}_{m, \mu}(a) = \text{arg}\max_{p} \left \Vert f_{m, \mu}(p) - f_{m, \mu}(a) \right \Vert_{2}^{2}$ and $\widehat{n}_{m, \mu}(a) = \text{arg}\min_{n} \left \Vert f_{m, \mu}(n) - f_{m, \mu}(a) \right \Vert_{2}^{2}$ respectively (Steps 2 and 3 of Fig. \ref{fig:system}). 
Now, combining these all together, we can formulate the final loss each adversary VAE trains on as
\begin{equation}
\begin{split}
    &\mathcal{L}_m^{\text{final}}(X_m^{(i)}, \bar{X}_m^{(i)}, z_m^{(i)}) = \mathcal{L}_m^{\text{VAE}}\left(X_m^{(i)}, \bar{X}_m^{(i)}, z_m^{(i)}\right) \\
    &+ \widehat{\lambda} \cdot \mathcal{L}_m^{\mu}\left(X_m^{(i)}, \widehat{p}_{m, \mu} (X_m^{(i)}), \widehat{n}_{m, \mu}(X_m^{(i)})\right), \label{eq:9}
\end{split}
\end{equation}
which is shown as Step 4 in Fig. \ref{fig:system}. Our experiments in Sec.~\ref{sec:simulations} will demonstrate the benefit of this hybrid VAE and metric learning approach, i.e., training on $\mathcal{L}_m^{\text{final}}$ versus $\mathcal{L}_m^{\text{VAE}}$.


After training the VAE, we introduce an auxiliary classifier $\phi_{m, \mu}$, which is trained in a supervised manner using the latent embeddings from the $\mu$ vector (Step 5 of Fig. \ref{fig:system}), denoted by $f_{m, \mu}(\cdot)$. 
We use the cross-entropy loss $\mathcal{L}(y, \hat{y}) = - \sum_{c \in \mathcal{C}} y_{c} \log(\hat{y}_c)$ as the objective function, where $y_{c}$ is an indicator for whether the data point is from class $c$, $\hat{y}_c$ is the softmax probability for the $c^{th}$ class, and $y$, $\hat{y}$ are the corresponding vectors. This is trained on $\{f_{m, \mu}(X_m^{(i)}) | X_m^{(i)} \in \widehat{\mathcal{D}}_m\}$.
In this way, the embeddings are trained to be separable, associating label positions in the latent space with their corresponding labels \cite{chen2020simple}. We can then employ this to construct the set of locally inferred target datapoints from adversary $m$, $\mathcal{D}_m^{(p)}$ (Step 6 of Fig.~\ref{fig:system}), as will be described in Sec.~\ref{sec:method-label}.


\subsubsection{\textbf{Collaborative Inference}}
Upon completing the local inference phase, the adversaries utilize their locally inferred labels to reach a consensus over the local graph $G$ on which datapoints are from the target label (Collaborative Inference phase of Fig. \ref{fig:system}).
This consensus can be reached in several potential ways, e.g., through a leader adversary node performing a Breadth-First Search (BFS) \cite{bellman1958routing} traversal on the graph, followed by a majority voting scheme (Step 7 of Fig. \ref{fig:system}). This is the method we will employ in Sec. \ref{sec:method-label}.

Each of these label inference processes, e.g., feature sharing, VAE training/inference, and collaborative inference, take place before the VFL training process begins. This emphasizes one of our contributions: developing a backdoor attack where gradients from the server are not utilized for label inference. As, we will see in Sec.~\ref{ssec:compAn}, this enhances robustness against server-side gradient noise injection defenses. Further, by minimizing the degree to which adversary behavior will deviate from the VFL training protocol, the possibility of the system detecting the attack (through e.g., anomalous communication, computation, or energy consumption) becomes small.


\subsection{Trigger Embedding}
\label{sec:backdoor}
In the next step, after the target data points have been inferred, the VFL training process starts, during which adversaries embed triggers into inferred data points. 
One of the primary challenges for the adversaries conducting a backdoor attack is to ensure that both the primary task and the backdoor task perform effectively.
To this end, adversaries poison a specific subset of the inferred data points, defined as $\mathcal{D}^{\text{pois}}$, according to a poisoning budget $\zeta = \frac{|\mathcal{D}^{\text{pois}}|}{|\mathcal{D}|}$. A smaller budget also helps prevent detection of the malicious operation.

Formally, during training iteration $t$, if any datapoint $i$ from $\mathcal{D}^{\text{pois}}$ is present in the minibatch $\mathcal{B}^{(t)}$, adversary $m$ will implant a trigger $\sigma_m$ into its local $x_m^{(i)}$ as $\widehat{x}_{m}^{(i)} = x_{m}^{(i)} + \sigma_{m}$.
Here, $\widehat{x}_{m}^{(i)}$ is the datapoint embedded with the trigger, and the adversary aims for the server to learn and associate this trigger pattern with the target label.
Unlike most existing works, our setup considers more than one adversary. Therefore, in Sec.~\ref{sec:trigger}, we will propose two different methods for generating $\sigma_m$ across adversaries $m \in \mathcal{A}$ as a subpartition or smaller version of the trigger $\sigma$ that would be embedded by a single adversary.

Our aim is to make the server rely on the trigger while still learning features relevant to the target label, by leveraging the VAE's generative properties. This involves the following steps:
\begin{itemize}[leftmargin=4mm]
    \item {\bf Data Generation:} Using adversary $m$'s VAE, we generate datapoints $\widetilde{x}_{m}^{(i)}$ by sampling vector $z_{m}^{\text{gen}} \sim \mathcal{N}(0, I)$ from the latent space and passing them through the decoder $p_{m, d}$. 
    \item {\bf Data Substitution and Selective Poisoning:} The adversary swaps the original datapoints $x_{m}^{(i)}$, similar to \cite{xuan2023practical}, with the newly generated ones $\widetilde{x}_{m}^{(i)}$, and embeds them with the trigger. This is performed on a subset of the inferred data points during training, according to the poisoning budget $\zeta$. 
\end{itemize}
As a result, the server learns more variations of the target label, which intuitively leads it to rely more on the trigger for classification.
Additionally, the generated samples will still follow the general structure of the target label, 
to prevent misclassification of labels not involved in the backdoor attack.

In designing the trigger, we aim for the poisoned datapoints to produce embeddings that are as close as possible to embeddings of non-poisoned embeddings. This can be thought of mathematically as looking for triggers that will minimize $\Vert \mathbb{E}[h_{m}^{\text{pois}}] - \mathbb{E}[h_{m}^{\text{target}}] \Vert_{2}$, where $\mathbb{E}[h_{m}^{\text{pois}}]$ is the expectation over feature embeddings produced from datapoints $i$ implanted with the trigger and $\mathbb{E}[h_{m}^{\text{target}}]$ is the expectation over feature embeddings produced from clean datapoints $i'$ belonging to the target label. 
If adversary $m$ had access to the server's loss function, it may be possible to incorporate (an approximation of) this norm difference directly into the adversary's local VFL update. However, we are considering a black-box scenario. Hence, to emulate the desired trigger behavior,
we complement the data substitution and selective poisoning process with an \textit{intensity-based triggering scheme}. Detailed in Sec.~\ref{sec:trigger}, this scheme enhances the background value of the trigger by an adversary-defined intensity value $\gamma$, so that the trigger becomes more prominent within the datapoint.
Thus, on the one hand, $\widetilde{x}_m^{(i)}$ presents the server with more variations of the target label, causing the server to rely more heavily on the consistent trigger pattern. Then, combining these harder-to-identify background features with the $\gamma$-enhanced trigger induces heavier reliance of the server on the trigger while minimizing main task performance degradation (when the trigger is not present).

The VFL system relies on the bottom models' outputted latent embeddings for classification, with updates to these local models aiming to optimize performance. By embedding the target datapoints with an intensified trigger, we can pull the embeddings $h_{m}^{\text{pois}}$ of the backdoored samples closer to embeddings $h_{m}^{\text{target}}$ of the target label, making it harder for the server to distinguish between them.
\section{Algorithm Details}
\label{sec:algorithm}

We now provide specific algorithms for implementing the label inference (Sec.~\ref{sec:method-label}) and trigger embedding (Sec.~\ref{sec:trigger}) methodologies from Sec.~\ref{sec:methodology}.

\subsection{Label Inference}
\label{sec:method-label}
Alg. \ref{algo:inference} summarizes our label inference approach. Recall the goal is to find datapoints of the target label $y_t$, i.e., to find which datapoints are candidates to be poisoned. This trains the VAE model $\phi_{m}^{\text{VAE}}$ and auxiliary classifier $\phi_{m, \mu}$ for adversary $m$, which is then capable of generating datapoints of $y_t$ based off local inference datapoints $\mathcal{D}_{m}^{(p)}$. As input, the overall algorithm will utilize the set of adversaries $\mathcal{A}$ and concatenated datapoints $\mathcal{D}_m$ for adversary $m$ to infer datapoints of $y_t$. We detail the steps of Alg. \ref{algo:inference} below:


\subsubsection{\textbf{Training VAE $\phi_{m}^{\text{VAE}}$ and Auxiliary Classifier $\phi_{m, \mu}$}}
As outlined in Sec. \ref{sec:inference}, label inference is performed before the standard VFL protocol. Firstly, $\phi_{m}^{\text{VAE}}$ is trained via \eqref{eq:9} and utilizes the ``batch-hard" strategy discussed in Sec. \ref{sec:inference} (Line 1 of Alg. \ref{algo:inference}). The VAE's reconstruction loss $\mathcal{L}_m^{\text{VAE}}$ employs  $\widehat{\mathcal{D}}_m^{\text{target}} \subseteq \widehat{\mathcal{D}}_m$, where $\widehat{\mathcal{D}}_m^{\text{target}}$ is the set of known concatenated datapoints belonging only to the target $y_t$. For the triplet loss $\mathcal{L}_m^{\mu}$, positives are selected from the VAE's training batch and the negatives are taken from $\widehat{\mathcal{D}}_m$. 
Next, $\phi_{m, \mu}$ takes in the $\mu$ embeddings on labeled datapoints and trains via cross-entropy loss (Line 2 of Alg. \ref{algo:inference}). 

After training the VAE, $\phi_{m, \mu}$ takes in the $\mu$-embeddings of the datapoints from $\mathcal{D}_m$ with unknown labels as input (Line 3 of Alg. \ref{algo:inference}), and populate $\mathcal{D}_{m}^{(p)}$, the set of locally inferred target datapoints. $\mathcal{D}_{m}^{(p)}$ is initially the same as $\widehat{\mathcal{D}}_m^{\text{target}}$ in terms of indices. The datapoints are added to the set if the maximum prediction probability corresponds to the target label $y_t$ and is greater than a confidence threshold $\beta$, i.e., $\max(\hat{y}) \geq \beta$ and $y_t = \text{arg}\max_c (\hat{y}_c)$ (Lines 5-7 of Alg. \ref{algo:inference}). In this way, only datapoints $\phi_{m, \mu}$ is confident about will be considered as targets.

Afterwards, adversary $m$'s VAE is retrained on $\mathcal{D}_{m}^{(p)}$ (Line 8 of Alg. \ref{algo:inference}). The retraining allows the target datapoint generation to match the shape of the feature partition.

\vspace{0.05 cm} 
\subsubsection{\textbf{Consensus Amongst Adversaries}}
After local label inference, collaborative inference begins (Line 9 of Alg. \ref{algo:inference}) with the following steps:
\begin{itemize}[leftmargin=4mm]
    \item {\bf Choosing Leader Node and BFS-traversal:} One adversary in graph $G$, denoted ${m}^{\text{lead}}$, is chosen by selecting the node with the highest degree. ${m}^{\text{lead}}$ will conduct BFS over $G$ to collect the local inference results from each adversary.
    \item {\bf Consensus Voting:} Next, given a multiset of the local inferred datasets {\small $\mathcal{D}^{(p)} = \{\mathcal{D}_{m}^{(p)}\}_{m \in \mathcal{A}}$}, ${m}^{\text{lead}}$ adopts a simple majority based voting scheme similar to \cite{naseri2023badvfl}. If an index $j$ appears for more than {\small$\left\lceil \frac{|\mathcal{A}|}{2} \right\rceil$} times in {\small$\mathcal{D}^{(p)}$}, it is added to the collaborative inference set {\small$\mathcal{D}_{g}^{(p)}$}. In other words,
    {\small$\mathcal{D}_{g}^{(p)} = \left\{ j \in \mathcal{D}^{(p)} \mid m(j, \mathcal{D}^{(p)}) > \left\lceil \frac{|\mathcal{A}|}{2} \right\rceil \right\}$}, 
    where $m(j, \mathcal{D}^{(p)})$ is the multiplicity of $j$ in $\mathcal{D}^{(p)}$ for an index $j$.
    \item {\bf Sharing Final Results:} Lastly, ${m}^{\text{lead}}$ will conduct BFS again, propagating the final indices of $\mathcal{D}^{(p)}$ to all adversaries, with $m$'s feature partition slice of $\mathcal{D}_{g}^{(p)}$ is defined as $\widetilde{\mathcal{D}}_m^{(p)}$.
\end{itemize}
\begin{figure}[t] 
    \centering
        \includegraphics[width=\linewidth]{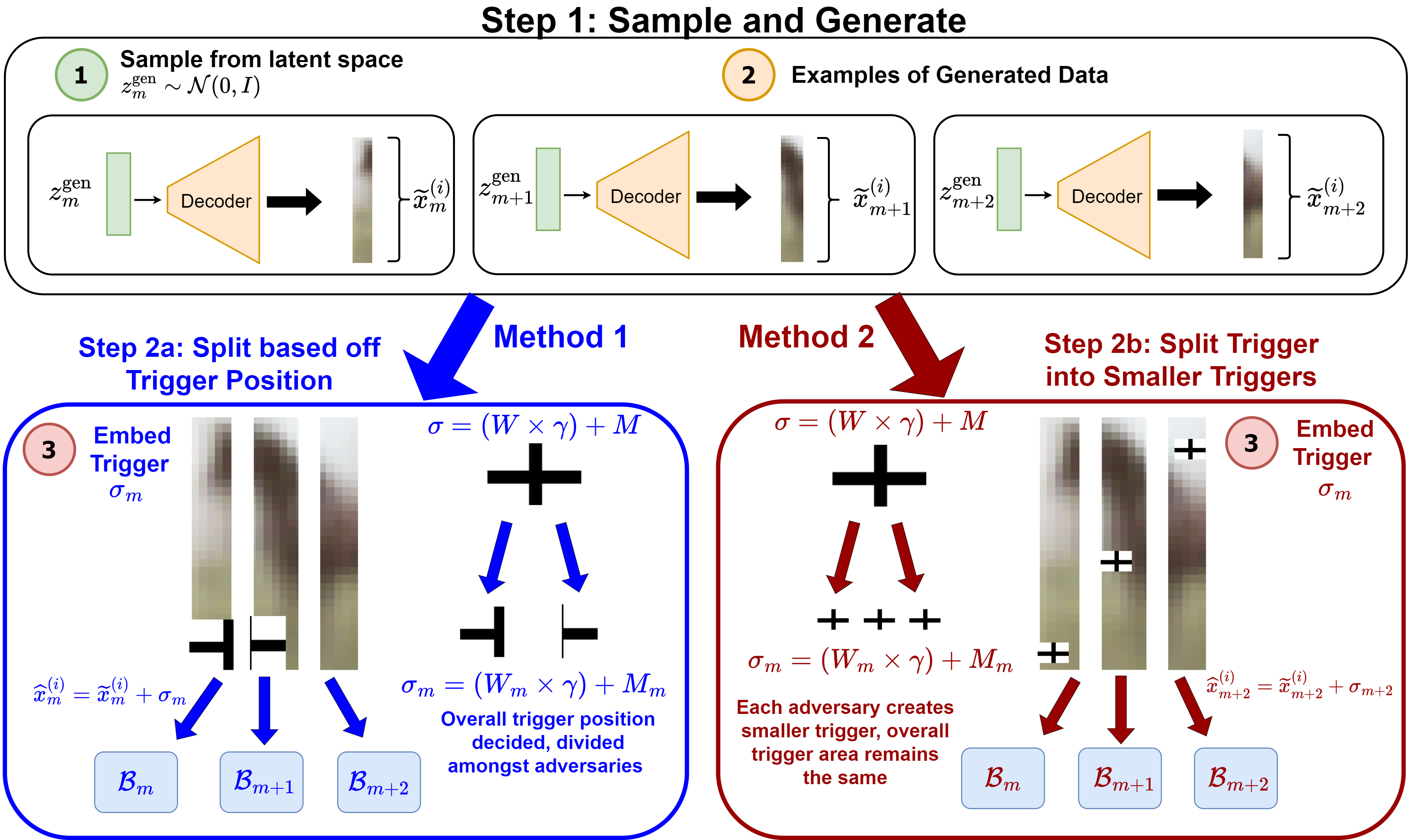}
        \vspace{-5mm}
        \caption{Image generation and trigger-embedding process. The adversaries can choose one of two methods: (1) constructing a collaborative trigger on some position of the known adversary features, or (2) giving each adversary a smaller trigger. Method 1 may result in some adversaries not possessing any portion of the trigger pattern, i.e., only having the background.}
        \label{fig:adver}
        \vspace{0.4 mm} 
\end{figure}
While we adopt the BFS traversal-based method, other graph traversal methods (e.g., DFS, spanning tree) could also be used, as long as all adversaries correctly receive the 
voting results.

\setlength{\textfloatsep}{0pt}
\begin{algorithm}[t]
{\small
\caption{VAE and Metric-Based Label Inference} 
\label{algo:inference}
\KwInput{VAE model $\phi_{m}^{\text{VAE}}$, auxiliary model $\phi_{m, \mu}$, adversary index $m \in \mathcal{A}$, target $y_t$, known concatenated datapoints $\widehat{\mathcal{D}}_m$, all concatenated datapoints $\mathcal{D}_m$}
Train $\phi_{m}^{\text{VAE}}$ using $\mathcal{L}_m^{\text{final}}$ from \eqref{eq:9}, which uses \eqref{eq:5} and \eqref{eq:6} \\
Train auxiliary classifier $\phi_{m, \mu}$ via $\{f_{m, \mu}(X_m^{(i)}) | X_m^{(i)} \in \widehat{\mathcal{D}}_m\}$ \\
\For {adversary index $m \in \mathcal{A}$}{
\For {datapoint $X_m^{(i)} \notin \widehat{\mathcal{D}}_m$ \textbf{and} $X_m^{(i)} \in \mathcal{D}_m$}{
     $\hat{y} = \phi_{m, \mu}(f_{m, \mu}(X_m^{(i)}))$ \\
    \If{$y_t = \text{arg}\max_c (\hat{y}_c)$ \textbf{and} $\max(\hat{y}) \geq \beta$}{
         Insert $x_m^{(i)}$ to $\mathcal{D}_{m}^{(p)}$
    }
}
Adjust model architecture and retrain $\phi_{m}^{\text{VAE}}$ on $\mathcal{D}_{m}^{(p)}$
}
Get indices $\mathcal{D}_{g}^{(p)} \leftarrow \text{CONSENSUS}(\mathcal{A})$ from Sec. \ref{sec:method-label} \\
\For {$j \in \mathcal{D}_{g}^{(p)}$}{
    Insert datapoint $x_m^{(j)}$ to $\widetilde{\mathcal{D}}_{m}^{(p)}$
}
\KwOutput{VAE $\phi_{m}^{\text{VAE}}$, inferred adversary dataset $\widetilde{\mathcal{D}}_{m}^{(p)}$}
}
\end{algorithm}

\subsection{Trigger Embedding}
\label{sec:trigger}
Alg. \ref{algo:trigger} summarizes our trigger embedding algorithm. Recall that after label inference, our goal is to conduct trigger embedding on the inferred datapoints. Overall, during the VAE protocol, we implant a trigger on generated datapoints from VAE $\phi_m^{\text{VAE}}$. We detail the steps of Alg. \ref{algo:trigger} below:


\subsubsection{\textbf{Poisoning and Trigger Implantation}}
To maintain a high accuracy in the main task along with the backdoor, a poisoning budget $\zeta$ outlined in Sec. \ref{sec:backdoor} is utilized to limit the number of poisoned datapoints. The selected indices to be poisoned are chosen at random from $\mathcal{D}_{g}^{(p)}$, with the corresponding sub-dataset for adversary $m$ denoted $\mathcal{D}_{m}^{\text{pois}} \subset \mathcal{D}^{\text{pois}}$. 

Now, the trigger implantation follows a two step process, depicted in Fig. \ref{fig:adver}. Firstly, a to-be-poisoned datapoint in a minibatch is replaced with a datapoint $\widetilde{x}_{m}^{(i)}$ generated from the decoder of the VAE, $p_{m, d}$ (Line 6 of Alg. \ref{algo:trigger}). 
Secondly, an intensity-based trigger is formed and distributed among adversaries. This can follow one of two methods, which we consider in the context of image data, where each sample $x$ is a matrix of pixels (or a tensor in the multi-channel case). Starting with Method 1, adversaries collaboratively insert a trigger into a target datapoint at a location specified by centering parameter \(\ell = (\ell_x, \ell_y)\). The trigger has a background of 1's with area \(W = h \times w\), divided among adversaries based on their feature partitions' proximity to \(\ell\). The background's value is enhanced by multiplying pixel-wise with an intensity parameter \(\gamma\), which controls the trigger's prominence. A cross pattern of 0's is added to complete the trigger (see Fig. \ref{fig:adver}). Overall, the general trigger pattern is similar to the trigger adopted in \cite{naseri2023badvfl}. Each adversary \(m\) receives a portion of the trigger according to their position relative to \(\ell\). The final datapoint with backdoor implantation is given by
\begin{equation}
    \widehat{x}_{m}^{(i)} = \widetilde{x}_{m}^{(i)} + ((W_{m} \times \gamma) + M_{m}). \label{eq:14}
\end{equation}
Here, \(\widehat{x}_{m}^{(i)}\) is created by implanting the trigger into \(\widetilde{x}_{m}^{(i)}\) (Line 7 of Alg. \ref{algo:trigger}), where \(W_{m}\) is the trigger background portion falling within adversary $m$'s local partition, and \(M_{m}\) is the local cross pattern. The trigger pattern is limited by a maximum area budget \(\epsilon\) to avoid server detection, i.e., \(h \cdot w \leq \epsilon\).

\subsubsection{\textbf{Alternative Trigger Embedding Method}}
Next, we describe an alternative method for adversaries to implant a trigger, referred to as Method 2 in Fig. \ref{fig:adver}.
In this case, instead of one collaborative trigger, each adversary implants a smaller subtrigger on their feature partition. 
The smaller subtriggers, when combined, should still have the same total area as the collaborative trigger; for example, all of their individual areas can be 
{\small \(W_m = \frac{h \cdot w}{{\mathcal{|A|}}} \)}. 
The subtriggers are placed randomly within the datapoint, preventing the server from memorizing the trigger pattern by location to enhance generalizability.
\vspace{-3mm}

\begin{algorithm}[t]
\caption{Distributed Trigger Embedding} 
\label{algo:trigger}
{\small
\KwInput{Server $K$, client $k \in \mathcal{K} \setminus \{K\}$, adversary set $\mathcal{A}$, adversary VAE $\phi_{m}^{\text{VAE}}$, bottom-model parameters $\theta_{k}$, to-be-poisoned data $\mathcal{D}_{m}^{\text{pois}} $ when $m \in \mathcal{A}$}
\For{$t=0$ to $T\!-\!1$}{
 \parfor{clients $k \in \mathcal{K} \setminus \{K\}$}{
     Sample local minibatch $\mathcal{B}_{k}^{(t)}$ \\
    \For{datapoint $x_{k}^{(i)} \in \mathcal{B}_{k}^{(t)}$}{
        \If{$x_{k}^{(i)} \in \mathcal{D}_{m}^{\text{pois}}$ \textbf{and} client $k \in \mathcal{A}$}{
             Adversary generates $\widetilde{x}_{k}^{(i)}=p_{m, d}(z_k^{\text{gen}})$ \\
             Add trigger pattern from \eqref{eq:14} \\
             Adversary replaces original data: $x_{k}^{(i)} = \widehat{x}_{k}^{(i)}$
        }
 }
     Compute $H_k^{(t)} \!=\! \{h_k^{(i)} \!=\! f_k(x_k^{(i)};\theta_k^t) | x_{k}^{(i)} \in  \mathcal{B}_k^{(t)}\}$ \\
     Transmit $H_{k}^{(t)}$ to server $K$
}
 Server computes $\{ \frac{\partial \mathcal{L}}{\partial h_k^{(i)} } | h_k^{(i)} \in H_k^{(t)} \}, k \in \mathcal{K} \setminus \{K\}$, sends them back to client $k$, computes $\frac{\partial \mathcal{L}}{\partial \theta_{_K}}$, and updates $\theta_{_K}^{(t+1)} \leftarrow \theta_{_K}^{(t)} - \eta_{_K}^{(t)} \frac{\partial \mathcal{L}}{\partial \theta_{_K}}$ \\
 \parfor{each client $k \in \mathcal{K} \setminus \{K\}$}{
         Compute $\frac{\partial \mathcal{L}}{\partial \theta_{k}}$ and update  $\theta_{k}^{(t+1)} \leftarrow \theta_{k}^{(t)} - \eta_{k}^{(t)} \frac{\partial \mathcal{L}}{\partial \theta_{k}}$
  }
}}
\vspace{-1mm}
\end{algorithm}


\vspace{0.35 cm}
\section{Convergence Analysis}
\label{theorem}
We now analyze the convergence of VFL in the presence of backdoor attacks. We first make the following assumptions:

\begin{assumption}[\textbf{$L$-Smoothness}]
\label{as:l-smooth}
The loss function \( F({\theta})\) in~\eqref{eq:1} is \(L\)-smooth, meaning that for any $x$ and $y$, we have
\[
F(y) \leq F(x) + \langle \nabla F(x), y - x \rangle + \frac{L}{2} \|y - x\|^2.
\]
\end{assumption}

\begin{assumption}[\textbf{Variance}]
    \label{as:variance_grad}
The mini-batch gradient $\nabla_{\theta_{k}}  \mathcal{L}$ is an unbiased estimate of $\nabla_{\theta_{k}}  F({\theta})$, and 
\[
\mathbb{E} \| \nabla_{\theta_{k}}  \mathcal{L} - \nabla_{\theta_{k}}  F({\theta}) \|^2 \leq \Gamma, \forall k \in \mathcal{K},
\]
where $\Gamma$ is the variance of $\nabla_{\theta_{k}}  \mathcal{L}$.
\end{assumption}

\begin{assumption}[\textbf{Perturbation}]
\label{as:bounded-grads} 
There is an upper bound for the gradient perturbation from adversaries, i.e., 
\[
\mathbb{E}\|\nabla^a_{\theta_{k}} \mathcal{L} - \nabla_{\theta_{k}}  \mathcal{L}\|^2 \leq \delta (\rho), \forall k \in \mathcal{K},
\]
where $\nabla^a_{\theta_{k}} \mathcal{L}$ 
denotes the perturbed gradient and $\rho$ is a measure of connectivity for the graph $G$.
\end{assumption}

Assumptions \ref{as:l-smooth} and \ref{as:variance_grad} are common in literature \cite{bottou2018optimization,wen2022communication}, while Assumption \ref{as:bounded-grads} characterizes the perturbation induced by the attack. We expect that $\delta (\rho)$ is an increasing function of $\rho$: a higher graph connectivity for collusion implies a higher probability that more datapoints will be targeted, leading to a larger perturbation. More specifically, as $\rho$ increases, adversary $m$'s concatenated sample $X_m^{(i)}$ becomes closer to the full feature set for each datapoint $i$. With more features, $m$ can infer more samples and achieve lower expected loss $\mathcal{L}(y, \hat{y})$ in local label inference. Consequently, adversaries will tend to perturb more samples from the target label, and the expected perturbation in Assumption \ref{as:bounded-grads} increases. Here, we can also consider a simple illustrative example: suppose there are multiple compromised nodes in an image classification task, aiming to perturb samples corresponding to the ``truck'' label. Each adversary holds a small portion of the overall features (blocks of pixels). When connectivity is low (e.g., as with a line graph topology), each adversarial node receives only a small share of features from neighbors. This may leave the adversaries without enough information after exchange to reliably identify samples containing a truck, such as when all of the received features correspond to pixels offset from the object (e.g., the surrounding sky). In contrast, with a higher $\rho$ (e.g., as with a fully connected graph topology), each adversary gains a more complete view of the contents of sample $i$ in its concatenated $X_m^{(i)}$. This gives them a higher chance of identifying presence of trucks, resulting in more samples to poison after the collaborative inference process, and thus a higher $\delta$. We will observe this relationship experimentally in Sec. \ref{sec:simulations}, where $\rho$ is taken as the second smallest eigenvalue of the Laplacian matrix of $G$ (the algebraic connectivity or Fiedler value) \cite{fiedler1973algebraic}, which often increases with average node degree.

$\delta$ is thus the parameter that connects the attack performance to the graph connectivity $\rho$. We demonstrate its impact on the model convergence in the following theorem:


\begin{theorem}
Suppose that the above assumptions hold, and the learning rate is upper bounded as $\eta_1^{(t)} = \eta_2^{(t)} = \cdots = \eta_{K}^{(t)} = \eta^{(t)} \leq \frac{1}{4L}$. Then, the iterates generated by the backdoored SplitNN and vanilla SplitNN satisfy
\begin{align}\label{VFL_bound}
  & \min_{t \in \{0, ..., T-1\}} \{ \mathbb{E} \|  \nabla F({\theta}^{(t)}) \|^2\} \!\leq  4\frac{F({\theta}^{(0)}) }{\sum_{t=0}^{T-1}\eta^{(t)} }   \nonumber  \\
  &\!+\! 4 \frac{\sum_{t=0}^{T-1}(\eta^{(t)})^2}{\sum_{t=0}^{T-1}\eta^{(t)}} (KL \Gamma + KL \delta (\rho)) +\! 2 K \delta (\rho).
\end{align}
\end{theorem}
\noindent \textbf{Proof:} The proof is contained in Appendix \ref{proof:theorem}.

When there is no attack, i.e., $\delta (\rho) = 0$, the bound in Theorem 1 recovers the result of VFL in \cite{chen2020vafl}. 
Under a learning rate $\eta^{(t)}$ that satisfies $\sum_{t=0}^{T-1}  (\eta^{(t)})^2 \rightarrow 0$ and $\sum_{t=0}^{T-1}  \eta^{(t)} \rightarrow \infty$ for $T\rightarrow \infty$, the first and second terms in the right-hand side of inequality \eqref{VFL_bound} diminish to zero.
The adversarial attack induces a constant term $2 K \delta (\rho)$ within the convergence bound, reflecting the convergence degradation due to the adversarial perturbations. 
Since $\delta (\rho)$ is an increasing function in terms of the connectivity, we see that the gap from a stationary point induced by the backdoor attack becomes progressively larger. 
\section{Numerical Experiments}
\label{sec:simulations}
In this section, we evaluate the performance of our proposed approach on various datasets. We compare our performance with two state-of-the-art backdoor VFL attack methods discussed in Sec.~\ref{ssec:related}: BadVFL \cite{xuan2023practical} and VILLAIN \cite{bai2023villain}.

\subsection{Simulation Setup}

\begin{table}[t]
 \vspace{2mm}
 \centering
 \footnotesize
 \setlength{\tabcolsep}{4pt} 
 \caption{\small Network architecture and hyperparameters for each dataset.\vspace{-.5mm}}
 \label{table:parameter}
 \begin{minipage}{0.75\linewidth}
     \centering
     \begin{tabular}{|c|c|c|}
        \hline
        \multirow{2}{*}{\textbf{Parameters}} & \multicolumn{2}{c|}{\textbf{Dataset}} \\ 
        \cline{2-3}
         & \textbf{MNIST} & \textbf{FMNIST} \\ 
        \hline
        Margin $\kappa$ & $0.4$ & $0.25$ \\ \hline
        Poisoning budget $\zeta$ & $1\%$ & $1\%$ \\ \hline
        Intensity factor $\gamma$ & $20$ & $30$ \\ \hline
        Confidence threshold $\beta$ & $0.999$ & $0.999$ \\ \hline
        $\#$ Auxiliary samples & $360$ & $360$ \\ \hline
        $\%$ Auxiliary as Target & $16\%$ & $16\%$ \\ \hline
        VAE Latent Dimension & $32$ & $64$ \\ \hline
     \end{tabular}
 \end{minipage}
 \vspace{3mm} 
 
 \begin{minipage}{0.95\linewidth}
     \centering
     \begin{tabular}{|c|c|c|c|}
        \hline
        \multirow{2}{*}{\textbf{Parameters}} & \multicolumn{3}{c|}{\textbf{Dataset}} \\ 
        \cline{2-4}
         & \textbf{CIFAR-10} & \textbf{SVHN} & \textbf{CIFAR100-20} \\ 
        \hline
        Margin $\kappa$ & $0.2$ & $0.2$ & $0.2$ \\ \hline
        Poisoning budget $\zeta$ & $1\%$ & $1\%$ & $0.5\%$ \\ \hline
        Intensity factor $\gamma$ & $30$ & $20$ & $95$ \\ \hline
        Confidence threshold $\beta$ & $0.9985$ & $0.99995$ & $0.99999$ \\ \hline
        $\#$ Auxiliary samples & $350$ & $560$ & $487$ \\ \hline
        $\%$ Auxiliary as Target & $14\%$ & $12.5\%$ & $15.4\%$ \\ \hline
        VAE Latent Dimension & $512$ & $256$ & $512$ \\ \hline
     \end{tabular}
 \end{minipage}
\end{table}

We perform experiments on the MNIST \cite{deng2012mnist}, Fashion-MNIST (FMNIST) \cite{xiao2017fashion}, CIFAR-10 \cite{Krizhevsky09learningmultiple}, Street View House Numbers (SVHN) \cite{netzer2011reading}, and CIFAR-100 (CIFAR100-20, coarse label version) \cite{Krizhevsky09learningmultiple, patacchiola2020self, novack2023chils} datasets.
We consider fully-connected VAEs for MNIST and FMNIST, a Convolutional VAE (CVAE) with 4 layers each for the decoder and encoder for CIFAR-10 and CIFAR100-20, and a 4-layer encoder and 3-layer decoder CVAE for SVHN. For the bottom model, MNIST and FMNIST adopt a two layer Convolutional Neural Network (CNN), with CIFAR-10 and CIFAR100-20 having the same architecture as the encoder of the VAE. SVHN has the same bottom model as CIFAR-10. For the top model, we adopt a two-layer fully-connected network, which trains using the Adam optimizer \cite{kingma2014adam}. Further details are given in Table \ref{table:parameter}.

In our experiments, unless stated otherwise, there are 10 clients with 5 adversaries, with the adversaries utilizing trigger Method 1 from Sec.~\ref{sec:trigger} by default. Moreover, since the baselines do not consider adversary graphs in their methodology, we by default consider a fully-connected graph for fair comparison. In addition, note that both baselines assume only one adversary, and we extend their method to multiple adversaries by adding majority voting and trigger splitting to their label inference and attack processes. All experiments were conducted on a server with a 40GB NVIDIA A100-PCIE GPU and 128GB RAM, using PyTorch \cite{paszke2019pytorch} for neural network design and training.

\subsection{Competitive Analysis with Baselines}
\label{ssec:compAn}

\begin{figure*}[t!]
    \begin{center}
        \begin{subfigure}{0.99\textwidth}
        \includegraphics[width=0.99\linewidth]{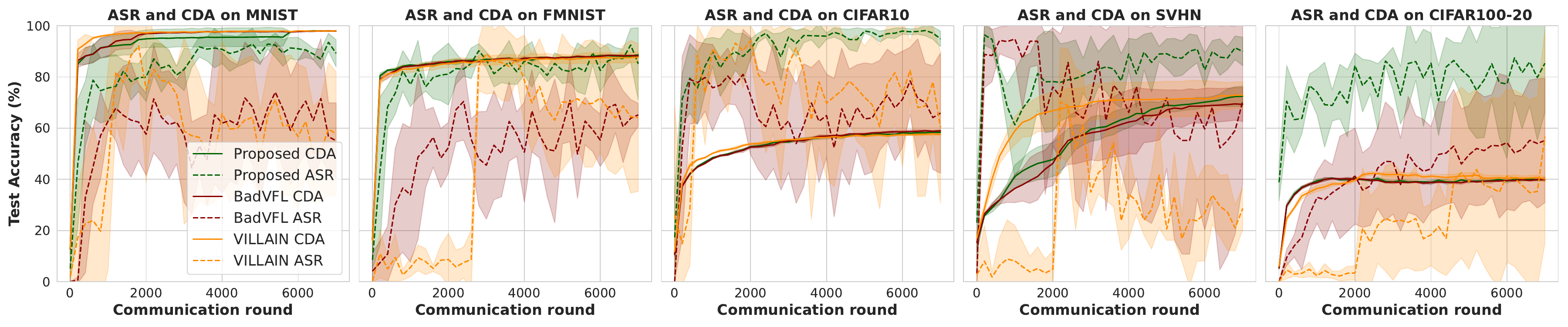}
        \end{subfigure}
    \end{center}
    \vspace{-4.5mm}
    \caption{Attack Success Rate (ASR) and Clean Data Accuracy (CDA) for MNIST, Fashion-MNIST, CIFAR-10, SVHN, and CIFAR100-20. The proposed method converges to a higher ASR value than the baselines (BadVFL \cite{xuan2023practical} and VILLAIN \cite{bai2023villain}) due to (1) having a higher label inference accuracy as seen in Fig. \ref{fig:label-inf} and (2) having an intensity based trigger that makes it easier for the server to draw the association between the target label and trigger.}
    \vspace{-0.12in} 
    \label{fig:asrcda}
\end{figure*}

\begin{figure*}[t!]
 \vspace{-2mm}
    \begin{center}
        \begin{subfigure}{0.24\textwidth} 
            \includegraphics[width=\linewidth]{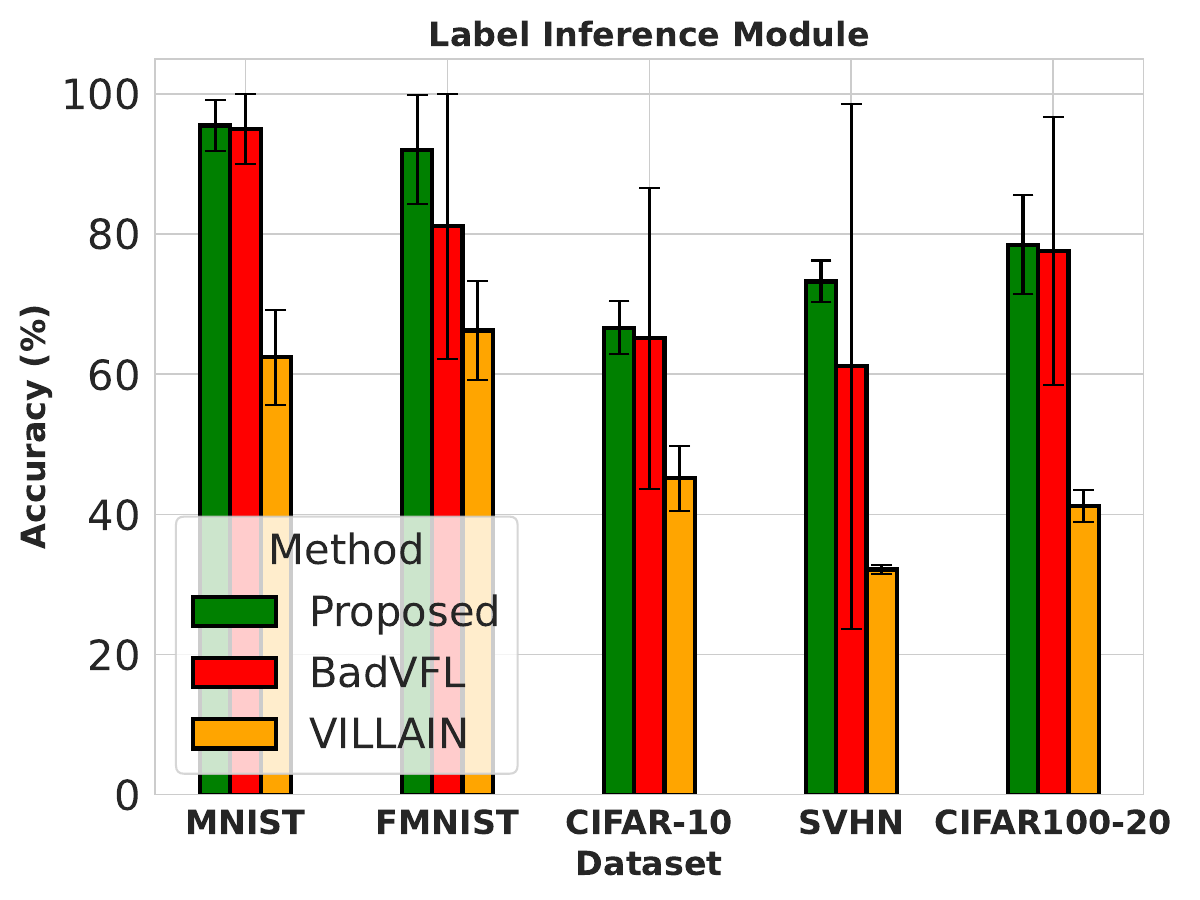}
            \caption{\small}
            \label{fig:label-inf}
        \end{subfigure}
        \begin{subfigure}{0.24\textwidth} 
            \includegraphics[width=\linewidth]{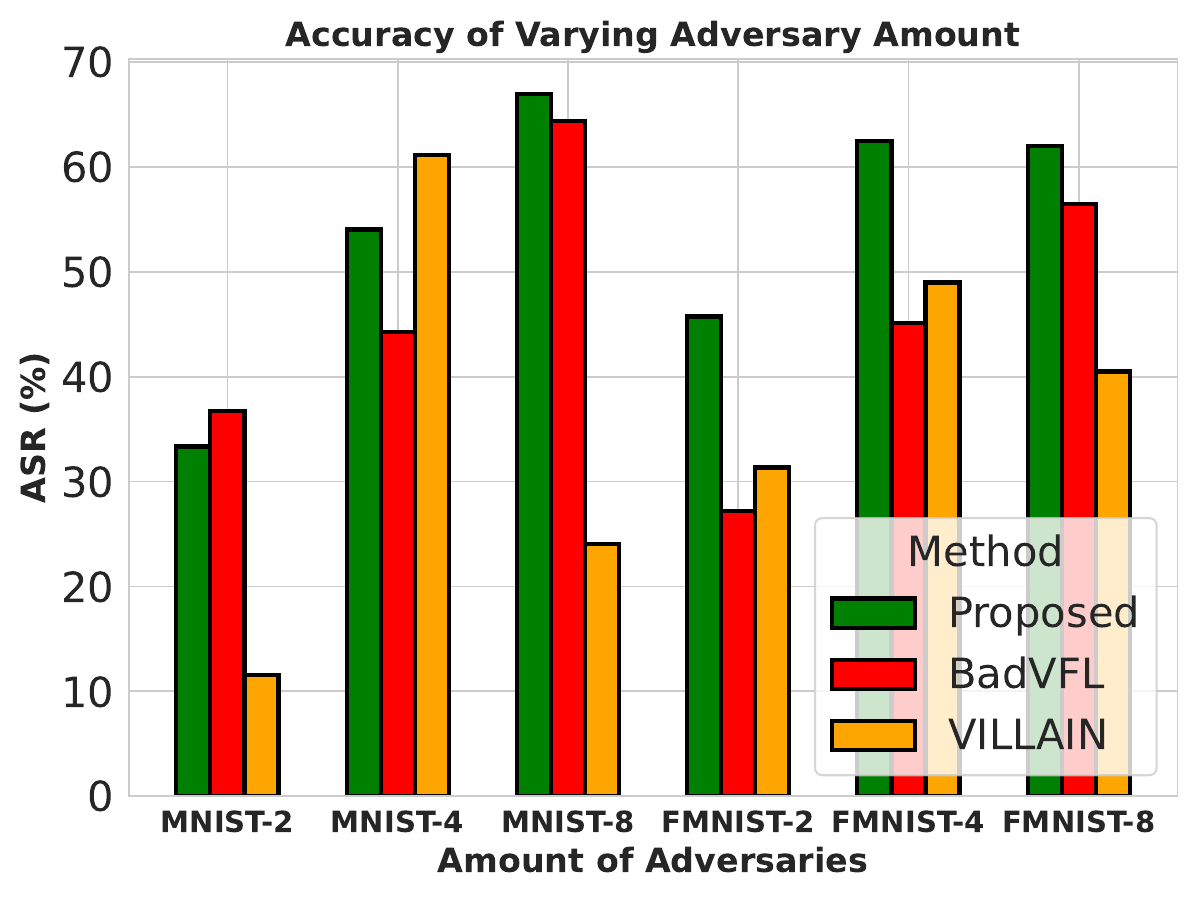}
            \caption{\small}
            \label{fig:varying-adversaries}
        \end{subfigure}
        \begin{subfigure}{0.255\textwidth} 
            \includegraphics[width=\linewidth]{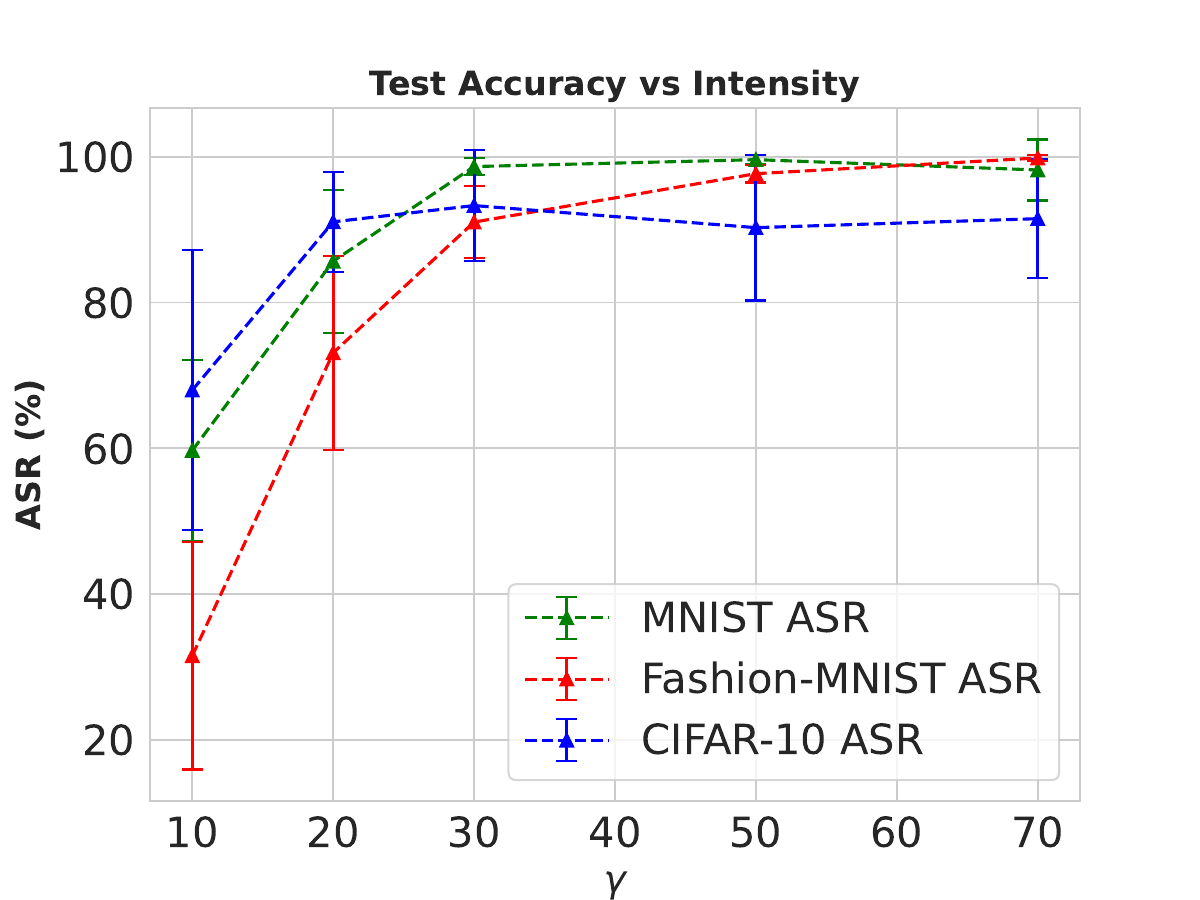}
            \caption{\small}
            \label{fig:intensity}
        \end{subfigure}
        \begin{subfigure}{0.24\textwidth}
            \includegraphics[width=\linewidth]{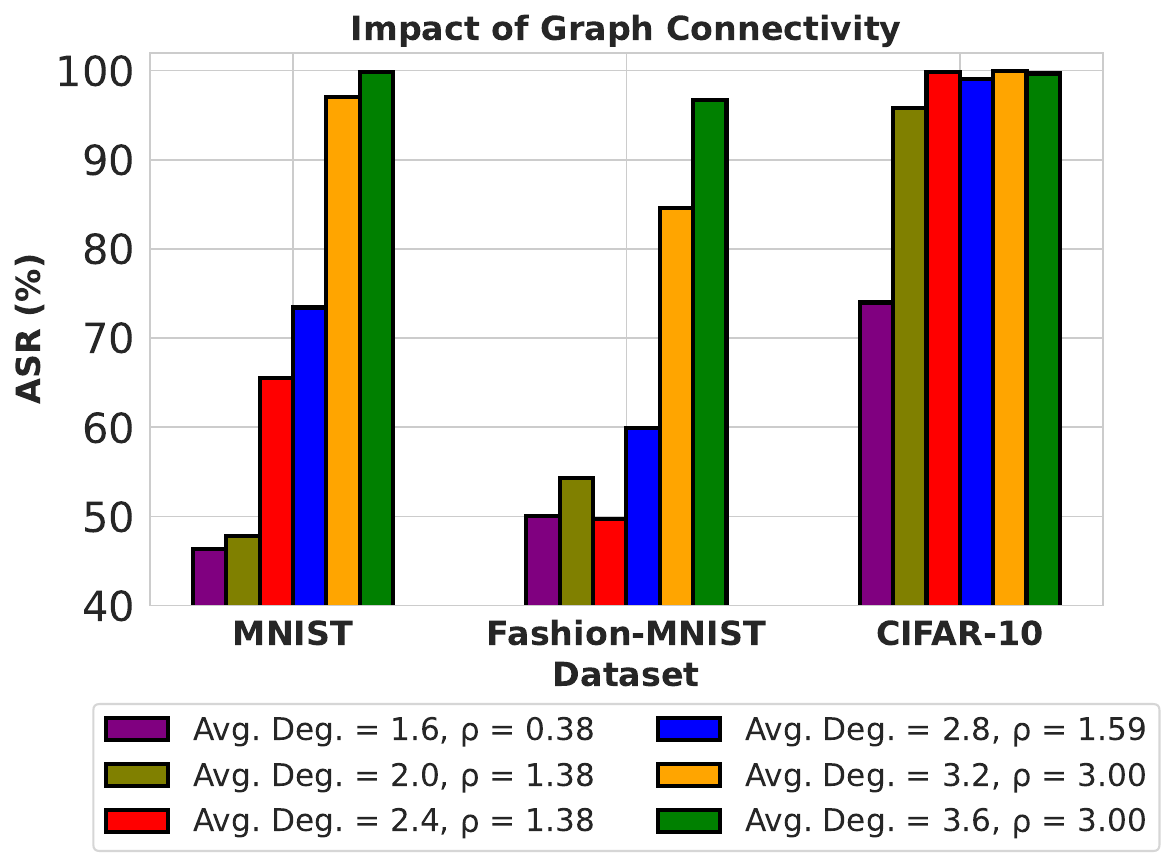}
            \caption{\small}
            \label{fig:graph-conn}
        \end{subfigure}
    \end{center}
    \vspace{-4mm}
    \caption{(a) The accuracy of label inference across all three scenarios. The proposed method reaches higher accuracy than the baselines (BadVFL \cite{xuan2023practical} and VILLAIN \cite{bai2023villain})  even without access to server information. (b) Impact of varying number of adversaries with Fashion-MNIST and MNIST. As the number of adversaries increases, a general trend in the increase of the ASR is noticed. In addition, the proposed attack is consistently has a higher or comparable ASR value to the baselines. (c) Varying trigger intensity parameter $\gamma$. We notice that an increase in $\gamma$ allows for a greater success of a backdoor attack. (d) Performance of the ASR with different levels of graph connectivity. In general, the more connected the graph is, the better the performance of the backdoor attack.}
    \vspace{-0.05in}
\end{figure*}

\begin{table*}[t!]
    \begin{center}
        \scriptsize
        \setlength{\tabcolsep}{3pt} 
        \caption{Impact of adding Gaussian noise to the gradients as a server-side defense: our proposed method is much more resistant to the defense due to not relying on the server for label information. The baselines, particularly VILLAIN, experience a significant drop in ASR when noise is introduced to the gradients.\vspace{-2mm}}
        \label{table:defense}
        \begin{tabular}{|c|c|c|c|c|c|c|c|c|c|}
        \hline
        \multirow{2}{*}{\textbf{Task}}  & \multicolumn{3}{c|}{\textbf{MNIST}} & \multicolumn{3}{c|}{\textbf{FMNIST}} & \multicolumn{3}{c|}{\textbf{CIFAR-10}} \\
        \cline{2-10}
         & Proposed & BadVFL \cite{xuan2023practical} & VILLAIN \cite{bai2023villain} & Proposed & BadVFL \cite{xuan2023practical}& VILLAIN \cite{bai2023villain} & Proposed & BadVFL \cite{xuan2023practical}& VILLAIN \cite{bai2023villain} \\ 
        \hline
        CDA          & $97.96 \pm 0.18$ & $97.91 \pm 0.15$ & $97.87 \pm 0.07$  & $88.54 \pm 0.08$ & $88.56 \pm 0.18$ & $88.86 \pm 0.11$   & $58.24 \pm 0.5$ & $58.62 \pm 0.46$ & $58.06 \pm 0.32$ \\ 
        CDA w/ Noise & $97.97 \pm 0.19$ & $97.83 \pm 0.2$ & $97.51 \pm 0.15$   & $88.25 \pm 0.41$ & $88.07 \pm 0.43$ & $85.98 \pm 3.0$    & $47.84 \pm 3.63$  & $37.12 \pm 14.18$ & $33.25 \pm 13.48$ \\ 
        ASR          & $94.89 \pm 3.76$ & $36.53 \pm 35.9$ & $53.29 \pm 12.78$ & $88.23 \pm 4.12$ & $63.40 \pm 15.73$ & $41.57 \pm 21.69$ & $93.95 \pm 6.86$ & $36.45 \pm 35.8$ & $67.10 \pm 16.78$ \\ 
        ASR w/ Noise & $\mathbf{86.88} \pm 4.03$ & $25.88 \pm 16.97$ & $0.66 \pm 0.38$  & $\mathbf{81.47} \pm 7.26$ & $38.64 \pm 16.07$ & $1.33 \pm 1.33$   & $\mathbf{89.50} \pm 7.16$ & $9.38 \pm 15.23$  & $2.26 \pm 2.36$ \\ 
        \hline
        \end{tabular}
    \end{center}
  \vspace{-4mm}
\end{table*}

\vspace{-1 mm}
\textbf{\textit{Attack Success Rate (ASR) and Clean Data Acc. (CDA): }}
First, we assess the accuracy of the backdoor attack and the main task and compare the performance of our approach to the BadVFL and VILLAN baselines. We define the accuracy of misclassifying a poisoned datapoint as the target label as Attack Success Rate (ASR) and the accuracy of the regular main task as Clean Data Accuracy (CDA), which is the same notation utilized in \cite{bai2023villain}. We intend to show a high ASR, indicating a successful backdoor, while keeping the CDA close to the baselines, indicating the main task is not significantly affected.

To measure the accuracy of both the attack (ASR) and the main task (CDA), we evaluate the main task on the test dataset, and select 250 random datapoints from the test set in each communication round that do not belong to the target label to embed the trigger for evaluation of the ASR. 
While we use a trigger dimension $W = 5 \times 7$ for the proposed method, for BadVFL \cite{xuan2023practical}, we follow their method of a white square, and set the trigger size to $9 \times 9$. For VILLAIN \cite{bai2023villain}, the trigger embedding method involves poisoning the embedding vector instead of the datapoint, and we poison $35\%$ of the embedding. 


Our results are given in Fig. \ref{fig:asrcda}. We note the superiority of our approach compared to the baselines: our attack achieves higher ASR across all settings, while the CDA stays relatively constant with the CDAs of the baselines. There are three reasons for this. First, due to better label inference performance (outlined in following section), the adversaries are more accurately poisoning datapoints belonging to the target, making it more likely for the server to draw an association between the target label and the trigger. Secondly, the intensity-based triggers are
more easily captured by the bottom-models, making the server more reliant on it for classification. Third, the samples generated by the adversary VAEs follow the same general patterns and features of the target label, meaning that it is still learning the overall structure of the clean features properly when poisoned datapoints are present, keeping both the ASR and CDA high.
Note that our findings are consistent on each of the datasets; compared to MNIST, Fashion-MNIST, CIFAR-10, and SVHN, the CIFAR100-20 experiments validate our results on a larger and more complex dataset. This corroborates the efficacy of our proposed method in implanting a successful backdoor on a wide range of learning scenarios.

We also note that our method requires a smaller trigger area compared to BadVFL \cite{xuan2023practical} to successfully carry out the attack. Additionally, BadVFL is sensitive to its initial known datapoint, causing the backdoor to fail during some runs, accounting for the high variation in the averages. As seen in Fig. \ref{fig:asrcda}, the ASR average is low for CIFAR100-20 and SVHN due to a poor initial known datapoint being used often. On the other hand, VILLAIN \cite{bai2023villain} must wait for a period of time before label inference and attacking can begin, since the method requires a well-trained bottom and top model to carry out an attack. Lastly, the baselines, particularly VILLAIN, slightly suffer from catastrophic forgetting, where the backdoor task is slowly forgotten over continuous iterations \cite{liu2024beyond}, whereas, our proposed attack consistently performs well.

\textbf{\textit{Label Inference:}} 
We compare the accuracy of our label inference to the baselines \cite{xuan2023practical, bai2023villain} in Fig. \ref{fig:label-inf}. 
We see that the proposed method of utilizing a combination of metric learning and VAEs for label inference has the best performance in terms of label inference accuracy across the datasets, while not relying on the server information. 
Note that the proposed attack can still be quite successful (Fig. \ref{fig:asrcda}) even when the label inference accuracy is not very high (e.g., with the CIFAR-10 and SVHN datasets).
This indicates that some margin of error in label inference can be acceptable, provided that the target data points significantly outnumber the combined total of other labels. 
As in Fig. \ref{fig:asrcda}, the inclusion of these five datasets validates the label inference accuracy module of our methodology across a wide range of learning scenarios. Specifically, while our proposed method achieves high label inference accuracy on benchmark datasets such as CIFAR-10 and MNIST, the results on the CIFAR100-20 dataset further demonstrate its robustness and scalability on more complex and diverse settings.


\begin{table*}[t!]
\centering
\caption{Study on the effectiveness of the proposed VAE-based swapping method for trigger embedding: our proposed method is much more resistant to forgetting the backdoor task due to the swapping of clean features with similar but harder-to-distinguish VAE-generated features. Moreover, the CDA still remains high, even with the swapping, indicating that it is still learning the overall structure of the clean features properly. Lastly, comparing the accuracy of the no attack scenario, we note that the accuracy does not significantly differ from the CDA, indicating the VAE-based swapping method successfully keeps both the ASR and CDA high. (For the No Atk. column, no results are given for the VAE-based swap row as no attack takes place.)}
\label{table:ablation-swap}
\vspace{0.1cm}
{\scriptsize
\begin{tabular}{lcccccccc}
\toprule
\multirow{2}{*}{\textbf{Task}} & \multicolumn{2}{c}{\textbf{MNIST}} & \multicolumn{2}{c}{\textbf{FMNIST}} & \multicolumn{2}{c}{\textbf{CIFAR-10}} & \multicolumn{2}{c}{\textbf{SVHN}} \\
\cmidrule(lr){2-3} \cmidrule(lr){4-5} \cmidrule(lr){6-7} \cmidrule(lr){8-9}
 & \textbf{VAE-based Swap} & \textbf{No Swap} & \textbf{VAE-based Swap} & \textbf{No Swap} & \textbf{VAE-based Swap} & \textbf{No Swap} & \textbf{VAE-based Swap} & \textbf{No Swap} \\ 
\midrule
\textbf{ASR-1} & $\mathbf{93.35} \pm 6.91$ & $32.97 \pm 18.48$ & $\mathbf{82.44} \pm 4.01$ & $46.46 \pm 11.51$ & $\mathbf{97.65} \pm 2.49$ & $12.13 \pm 8.01$ & $\mathbf{91.97} \pm 3.42$ & $40.58 \pm 9.00$ \\
\textbf{CDA-1} & $97.94 \pm 0.12$ & $97.82 \pm 0.12$ & $88.39 \pm 0.58$ & $88.23 \pm 0.24$ & $58.59 \pm 0.17$ & $58.31 \pm 0.21$ & $70.23 \pm 2.79$ & $70.20 \pm 1.78$ \\
\textbf{ASR-2} & $\mathbf{73.67} \pm 33.50$ & $0.33 \pm 0.67$ & $\mathbf{78.61} \pm 22.33$ & $26.34 \pm 49.15$ & $\mathbf{94.95} \pm 8.12$ & $0.33 \pm 0.43$ & $\mathbf{81.53} \pm 17.22$ & $5.30 \pm 9.67$ \\
\textbf{CDA-2} & $98.01 \pm 0.19$ & $97.86 \pm 0.10$ & $88.51 \pm 0.13$ & $88.48 \pm 0.23$ & $58.45 \pm 0.57$ & $58.52 \pm 0.37$ & $66.65 \pm 3.68$ & $69.83 \pm 9.13$ \\
\textbf{No Atk.} & --- & $97.84 \pm 0.30$ & --- & $86.96 \pm 2.14$ & --- & $58.99 \pm 0.47$ & --- & $71.24 \pm 9.07$ \\
\bottomrule
\end{tabular}}
\end{table*}

\begin{table*}[t!]
\centering
\caption{Study on the effectiveness of the proposed attack in terms of ASR with and without the use of the consensus-voting system utilizing trigger embedding Method 2 from Fig. \ref{fig:adver}. Note that many of the voting column results are the same as ASR-2 and CDA-2 with VAE-based swapping in Table \ref{table:ablation-swap} due to both being run on the same settings and hyperparameters.}
\label{table:ablation-vote}
\vspace{0.1cm}
\scriptsize
\begin{tabular}{lcccccccc}
\toprule
\multirow{2}{*}{\textbf{Task}} & \multicolumn{2}{c}{\textbf{MNIST}} & \multicolumn{2}{c}{\textbf{FMNIST}} & \multicolumn{2}{c}{\textbf{CIFAR-10}} & \multicolumn{2}{c}{\textbf{SVHN}} \\
\cmidrule(lr){2-3} \cmidrule(lr){4-5} \cmidrule(lr){6-7} \cmidrule(lr){8-9}
 & \textbf{Voting} & \textbf{No Vote} & \textbf{Voting} & \textbf{No Vote} & \textbf{Voting} & \textbf{No Vote} & \textbf{Voting} & \textbf{No Vote} \\ 
\midrule
\textbf{ASR} & $\mathbf{73.67} \pm 33.50$ & $0.22 \pm 0.45$ & $\mathbf{78.61} \pm 22.33$ & $25.0 \pm 50.0$ & $\mathbf{98.79} \pm 1.48$ & $34.45 \pm 38.86$ & $\mathbf{74.00} \pm 48.72$ & $24.00 \pm 45.40$ \\
\textbf{CDA} & $98.01 \pm 0.19$ & $97.84 \pm 0.30$ & $88.51 \pm 0.13$ & $86.96 \pm 2.14$ & $58.70 \pm 0.94$ & $59.19 \pm 0.84$ & $66.28 \pm 3.57$ & $69.50 \pm 7.74$ \\
\bottomrule
\end{tabular}
\end{table*}

\textbf{\textit{Robustness against Defense:}}
We evaluate the effectiveness of our attack against traditional server-side defense mechanisms, specifically noising defenses \cite{xuan2023practical, zhu2019deep}. We insert noise with variance $10^{-10}$ in the gradients sent back by the server, and assess the corresponding ASR values averaged over several runs in Table \ref{table:defense}.
We note that the proposed method maintains relatively small degradation in ASR performance compared to those obtained without any defenses, indicating the method's robustness to such strategies.
By contrast, the ASR values drop significantly in presence of noise-injected gradients for the competing baselines. 
The reason behind this is that the baselines both rely on server gradients to construct the attack: noise addition affects the similarity comparison of the baselines, leading to poor label inference. 
Moreover, the server can employ noising without significantly affecting the CDA, suggesting that it can defend against baselines \cite{xuan2023practical, bai2023villain}, but remains vulnerable to our method.
Overall, these results validate the robustness of our label inference based on hybrid VAEs rather than server gradients, which is crucial to maintaining a good ASR in the presence of such defenses.

\vspace{-0.1in}
\subsection{Varying Adversaries and Attack Network}
\textbf{\textit{Impact of Varying Number of Adversaries:}} Now, we analyze the impact of the number of adversaries on the ASR on the server's top model. 
For this experiment, we utilize the trigger embedding method of splitting the trigger into smaller separate local subtriggers (i.e., Method 2 from Fig. \ref{fig:adver}). We assume that each adversary holds an $8 \times 2$ trigger when there are two adversaries, $4 \times 2$ triggers when there are four, and $2 \times 2$ triggers when there are eight, thus all have the same total area of $32$.

The results are given in Fig. \ref{fig:varying-adversaries}. We see a general trend where an increase in the number of adversaries leads to a higher ASR, suggesting that more adversaries result in a stronger attack, even when the area threshold parameter $\epsilon$ remains unchanged. 
We also observed (not shown) that varying the number of adversaries did not significantly affect the CDA, indicating that the presence of multiple adversaries does not impact the main task across different methods. 

\vspace{1 mm}
\textbf{\textit{Impact of Trigger Intensity:}}
We next analyze the impact of the intensity value $\gamma$ on the ASR. The results across three datasets are shown in \ref{fig:intensity}.
The analysis reveals that a stronger trigger produces a higher ASR, with the top-performing server model showing a stronger association to the target as the trigger's intensity increases. 
Notably, while MNIST and CIFAR-10 achieve relatively good performance even at lower $\gamma$ values, Fashion-MNIST requires a higher trigger intensity to attain desirable ASR accuracy. 
This is likely due to a significant portion of Fashion-MNIST datapoints (being covered in white), matching the background of the trigger, thus necessitating a stronger trigger to differentiate from the clean features.



\vspace{0.2 mm}
\textbf{\textit{Graph Connectivity: }}
Next, we investigate how the algebraic connectivity $\rho$ of the adversary graph affects the attack performance. We simulate this by progressively increasing the average degree of the graph, beginning from a line topology, and consider Method 2 from Fig. \ref{fig:adver} for trigger embedding. The results are given in Fig. \ref{fig:graph-conn}. 
We see that the attack efficacy tends to get enhanced (i.e., ASR increases) with the increase of $\rho$, indicating that $\delta(\rho)$ from Theorem 1 is increasing in $\rho$. This is because the adversaries receive a higher share of the features for higher $\rho$. Note that due to the increased complexity of features in CIFAR10, the server becomes more reliant on the trigger, meaning a lower $\rho$ is sufficient to achieve a high ASR.



\vspace{0.2 mm}
\textbf{\textit{Ablation Studies: }}
\begin{table}[t]
    \begin{center}
        \footnotesize
        \setlength{\tabcolsep}{4pt} 
        \caption{\small Study of the effectiveness of the attack in terms of ASR with and without the use of triplet loss on the adversarial VAEs.\vspace{-1.5mm}}
        \label{table:loss}
        \begin{tabular}{|c|c|c|c|}
        \hline
        \multirow{2}{*}{\textbf{Architecture}} & \multicolumn{3}{c|}{\textbf{Dataset}} \\ 
        \cline{2-4}
         & \textbf{MNIST} & \textbf{FMNIST} & \textbf{CIFAR-10} \\ 
        \hline
        VAE Only & $84.00 \pm 9.78$ & $89.10 \pm 10.81$ & $76.18 \pm 11.93$ \\ 
        Hybrid VAE \& Triplet & $\mathbf{89.23} \pm 6.09$ & $\mathbf{91.74} \pm 4.61$ & $\mathbf{92.72} \pm 9.38$ \\ 
        \hline
        \end{tabular}
       \vspace{-1mm}
    \end{center}
\end{table}
We conduct several ablation studies. First, we explore the impact of incorporating the triplet loss into the VAE for label inference in~\eqref{eq:9}, i.e., whether it results in a higher attack potency. The corresponding results are shown in Table \ref{table:loss}, which demonstrate significant improvement in ASR values for all three datasets compared to the VAE-only loss.
Additionally, we note that the CIFAR-10 dataset experiences a significantly larger enhancement compared to MNIST and Fashion-MNIST datasets in terms of ASR:
the datapoints in CIFAR10 involve a more complex structure, where triplet loss may play a crucial role in refining embedding quality, facilitating a more effective attack. 
Overall, our findings validate our choice of hybrid VAE and metric learning for improving the attack performance.

In addition, we investigate the addition of the VAE-generated datapoint swapping mechanism in (\ref{eq:14}) to investigate whether it results in better overall attack performance for both Methods 1 (ASR-1 and CDA-1) and 2 (ASR-2 and CDA-2) outlined in Fig. \ref{fig:adver}. For Method 2, each adversary adopts a subtrigger size of $4 \times 2$ for a total area of $40$. Given the results in Table \ref{table:ablation-swap}, we show that the VAE-based swapping of original samples with VAE-generated samples $\widetilde{x}_{m}^{(i)}$ (VAE-based Swap in Table \ref{table:ablation-swap}) plays a crucial role in the success of the attack for both Methods 1 and 2. When no swapping mechanism is utilized (No Swap in Table \ref{table:ablation-swap}), the ASR is significantly lower ($\approx$ 85$\%$ and above with CIFAR-10) compared to when VAE-based swapping takes place, resulting in a failed attack. We also note that Method 1 of trigger embedding is more stable in its attack performance, with less variation in the averages compared to the attack utilized by Method 2. This is because the location where each adversary places the subtrigger for Method 2 varies instead being centered like Method 1, making it harder for the top model to learn the pattern. This means that having a denser adversary graph allows for better attack facilitation, as having a complete graph allows for the usage of Method 1. However, both methods still overall achieve high attack success rates across all four datasets. Moreover, when comparing the CDA of the proposed method to when no attack takes place, the accuracies are similar, indicating that the VAE generated samples still allow the top model to learn the general clean features of the target label well in addition to the trigger.

Finally, we also investigate the importance of the addition of majority voting in the label inference module by comparing the proposed method to no voting taking place. When no vote takes place, each adversary embeds its local trigger partition based off its local label inference results. Looking at the results presented in Table \ref{table:ablation-vote}, we notice a significant decrease in ASR performance when only local voting is considered on each adversary device, highlighting the importance of the majority voting module in the overall effectiveness of the attack. This is because no consensus has been reached on which final datapoints to poison, and so often the trigger implanted for each sample is incomplete, making it harder for the top model to fully learn the trigger pattern. Moreover, this also means that it is often the case that only a portion of the poisoned adversary-owned features is swapped out with VAE-generated sample $\widetilde{x}_{m}^{(i)}$, making it less likely for the server to rely on the trigger for classification due to the presence of more unchanged features, thereby resulting in a failed attack.

\vspace{0.2 mm}
\textbf{\textit{Effectiveness of Attack under Differing Latent Sizes: }} We analyze whether the latent embedding size that the server requires each local bottom model to send affects the overall effectiveness of the proposed attack, as outlined in Table \ref{table:dimension-impact}. We note that no matter the size of the embedding required by the server, the attack in terms of the ASR remains high. This means that the proposed attack can accommodate and learn the trigger with both small and large latent space dimensions. Moreover, when an increase in the latent space dimension improves the overall main learning task (CDA), as is the case with SVHN, the ASR remains largely unchanged. Overall, this indicates the robustness of the proposed attack to varying embedding sizes required by the server.

\begin{table}[t] 
\centering
\caption{CDA and ASR evaluation metrics for differing latent space dimensions of the embeddings produced by the bottom models on each client. For both small and large latent embedding values, the trigger pattern is successfully injected into the learning process of the server's top model, resulting in a successful backdoor attack.}
\small
\begin{tabular}{c|l|c|c}
\toprule
\textbf{Dimension} & \textbf{Datasets} & \textbf{ASR} & \textbf{CDA} \\
\hline
\multirow{4}{*}{$32$} & MNIST & $89.23 \pm 6.09$ & $97.92 \pm 0.15$ \\
 & FMNIST & $96.34 \pm 2.32$ & $88.33 \pm 0.09$ \\
 & CIFAR-10 & $95.74 \pm 4.21$ & $56.37 \pm 0.31$ \\
 & SVHN & $94.07 \pm 5.98$ & $45.02 \pm 9.60$ \\
\hline
\multirow{4}{*}{$64$} & MNIST & $87.07 \pm 9.65$ & $98.09 \pm 0.12$ \\
 & FMNIST & $98.53 \pm 1.13$ & $88.33 \pm 0.28$ \\
 & CIFAR-10 & $97.55 \pm 2.43$ & $57.55 \pm 0.10$ \\
 & SVHN & $90.50 \pm 10.36$ & $52.29 \pm 13.63$ \\
\hline
\multirow{4}{*}{$128$} & MNIST & $86.49 \pm 7.28$ & $98.15 \pm 0.06$ \\
 & FMNIST & $99.65 \pm 0.22$ & $87.33 \pm 2.66$ \\
 & CIFAR-10 & $96.03 \pm 2.16$ & $57.44 \pm 0.72$ \\
 & SVHN & $86.96 \pm 7.87$ & $60.28 \pm 7.88$ \\
 \hline
\multirow{4}{*}{$256$} & MNIST & $86.93 \pm 12.40$ & $98.02 \pm 0.16$ \\
 & FMNIST & $97.70 \pm 3.15$ & $88.45 \pm 0.14$ \\
 & CIFAR-10 & $95.65 \pm 5.14$ & $58.24 \pm 0.44$ \\
 & SVHN & $87.33 \pm 10.63$ & $69.85 \pm 3.06$ \\
 \hline
\multirow{4}{*}{$512$} & MNIST & $95.02 \pm 2.69$ & $98.18 \pm 0.08$ \\
 & FMNIST & $98.10 \pm 0.53$ & $87.72 \pm 2.36$ \\
 & CIFAR-10 & $97.38 \pm 1.44$ & $59.50 \pm 0.71$ \\
 & SVHN & $93.29 \pm 4.37$ & $71.57 \pm 1.92$ \\
\bottomrule
\end{tabular}
\vspace{2mm}
\label{table:dimension-impact}
\end{table}

\vspace{0.2 mm}
\begin{table*}[t!]
\centering
\caption{CDA and ASR evaluation metrics on different $\widehat{\kappa}$ margin values during triplet margin loss \eqref{eq:6} for label inference with the adversarial VAEs. Overall, the attack in terms of the ASR remains high across all datasets, but some margin values perform better than others and are more stable in their performance. In addition, with the exception of SVHN, the CDA stays mostly the same and is unaffected by the margin value adopted by the adversaries.}
\label{table:margin-impact}
\vspace{0.1cm}
\scriptsize
\begin{tabular}{lcccccccc}
\toprule
\multirow{2}{*}{\textbf{Margin ($\widehat{\kappa}$)}} & \multicolumn{2}{c}{\textbf{MNIST}} & \multicolumn{2}{c}{\textbf{FMNIST}} & \multicolumn{2}{c}{\textbf{CIFAR-10}} & \multicolumn{2}{c}{\textbf{SVHN}} \\
\cmidrule(lr){2-3} \cmidrule(lr){4-5} \cmidrule(lr){6-7} \cmidrule(lr){8-9}
 & \textbf{ASR} & \textbf{CDA} & \textbf{ASR} & \textbf{CDA} & \textbf{ASR} & \textbf{CDA} & \textbf{ASR} & \textbf{CDA} \\ 
\midrule
\textbf{0.10} & $91.13 \pm 5.78$ & $97.94 \pm 0.10$ & $94.67 \pm 1.96$ & $88.34 \pm 0.53$ & $95.82 \pm 2.73$ & $58.13 \pm 0.32$ & $91.97 \pm 2.66$ & $65.83 \pm 14.49$ \\
\textbf{0.15} & $91.36 \pm 2.58$ & $97.93 \pm 0.10$ & $88.12 \pm 3.11$ & $88.41 \pm 0.42$ & $92.37 \pm 4.76$ & $58.70 \pm 0.18$ & $81.44 \pm 10.97$ & $69.50 \pm 3.64$ \\
\textbf{0.2} & $87.30 \pm 9.62$ & $97.91 \pm 0.17$ & $91.59 \pm 2.27$ & $88.23 \pm 0.50$ & $97.76 \pm 1.09$ & $57.98 \pm 0.61$ & $89.28 \pm 7.76$ & $68.48 \pm 1.16$ \\
\textbf{0.25} & $86.17 \pm 11.38$ & $98.05 \pm 0.05$ & $85.10 \pm 16.31$ & $88.63 \pm 0.17$ & $87.87 \pm 9.08$ & $58.49 \pm 0.43$ & $89.14 \pm 6.15$ & $71.33 \pm 5.90$ \\
\textbf{0.30} & $92.54 \pm 1.17$ & $97.89 \pm 0.12$ & $87.99 \pm 7.12$ & $88.85 \pm 0.17$ & $91.16 \pm 3.83$ & $58.67 \pm 0.17$ & $92.15 \pm 3.56$ & $73.81 \pm 3.06$ \\
\textbf{0.35} & $90.38 \pm 6.52$ & $98.02 \pm 0.22$ & $95.46 \pm 4.89$ & $86.23 \pm 3.35$ & $91.97 \pm 7.88$ & $58.89 \pm 0.96$ & $83.15 \pm 7.27$ & $72.97 \pm 5.87$ \\
\textbf{0.40} & $89.23 \pm 6.09$ & $97.92 \pm 0.15$ & $96.14 \pm 4.90$ & $87.89 \pm 0.35$ & $94.90 \pm 3.56$ & $58.54 \pm 0.20$ & $95.16 \pm 2.90$ & $64.69 \pm 5.31$ \\
\textbf{0.45} & $86.39 \pm 9.05$ & $97.98 \pm 0.12$ & $94.99 \pm 2.99$ & $88.24 \pm 0.28$ & $97.00 \pm 2.18$ & $58.46 \pm 0.53$ & $90.52 \pm 6.18$ & $69.20 \pm 6.50$ \\
\bottomrule
\end{tabular}
\end{table*}
\textbf{\textit{Effectiveness of Attack under Differing Margins: }} Next, we analyze whether the choice of the margin value $\widehat{\kappa}$ when training the triplet-based VAE affects the overall performance of the backdoor attack, as outlined in Table \ref{table:margin-impact}. We note that in all data sets, the choice of $\widehat{\kappa}$ does not significantly impact both the CDA and ASR. This could be attributed to the use of an online batch-hard triplet mining strategy during training (Sec. \ref{sec:inference}), meaning difficult negatives are always used in the training of the adversarial VAEs, no matter the margin value chosen. However, some margin values, such as 0.3 vs 0.15 for SVHN or 0.25 vs 0.40 for FMNIST, perform better than others. In addition, with the case of SVHN, we note that arbitrarily selecting too high a margin value results in a $\approx 9\%$ drop in CDA performance. This might happen if the latent space is not well-separated, leading to many samples of differing classes to be selected. This could possibly create a lack of consistency in the type of data points that are trigger-embedded, in addition to the types of samples the VAE generates, leading to a degradation of the CDA.  Overall, these findings indicate that while the margin is not a highly sensitive parameter that requires meticulous tuning, some consideration should be taken when selecting an ideal value for maximal attack potency and main task performance.

\vspace{0.2 mm}
\textbf{\textit{Function Analysis of $\delta (\rho)$: }} Finally, we examine the relationship between $\delta (\rho)$ and $\rho$ from our analysis in Sec. \ref{theorem}. We compute $\delta$ as the L2 norm difference between the gradients used to update the top model under attack and under benign conditions. To enable broader analysis of connectivity over a larger range of $\rho$, we resize the samples \cite{dosovitskiy2020image, tu2023muller, nguyen2023single, pan2022towards} to $64 \times 64$. This adjustment allows for accommodating 20 total clients, including 10 adversaries, thereby supporting a denser adversary graph. Starting with a line topology, we incrementally add edges until the network becomes fully connected. Each adversary adopts a subtrigger size of $2 \times 2$ with a total trigger area of $40$. For CIFAR-10 and SVHN, the VAE architecture is adjusted to a 2-layer encoder and decoder CVAE.

\begin{figure}[t!] 
    \centering 
        \includegraphics[width=0.95\linewidth]{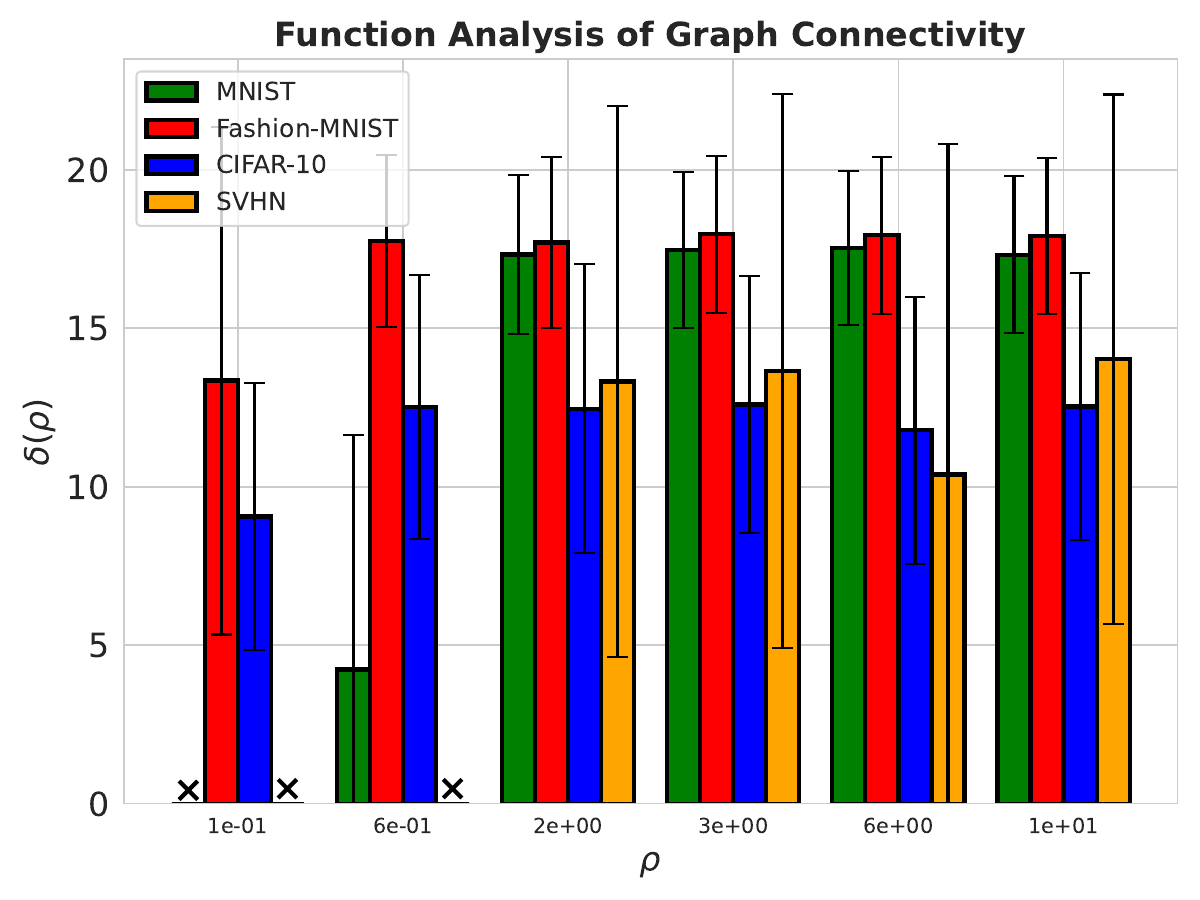}
        \vspace{-0.05in}
        \caption{\small The gradient perturbation in the top model from the adversaries for differing levels of connectivity $\rho$. This validates the hypothesis in Sec. \ref{theorem} that $\delta (\rho)$ is a non-decreasing function of $\rho$. As $\rho$ increases, the perturbation saturates quickly. When no perturbation takes place, it is marked with an ``x", as seen for some datasets when $\rho \in \{0.1, 0.6$\}.}
        \vspace{0.05in} 
        \label{fig:deltafunc}
\end{figure}

The results are shown for each dataset in Fig. \ref{fig:deltafunc}. We see that as the connectivity $\rho$ increases, the gradient perturbation introduced by the attack increases until it saturates once a certain level of connectivity is reached. When connectivity is lower, the number of poisoned datapoints is lower, resulting in a smaller disturbance during the training process. In addition, the poisoning budget $\zeta$ set by the adversaries prevents excessive poisoning, limiting the amount of perturbation presented to the top model. Moreover, we note that the perturbation increases more dramatically when the connectivity is lower, as individual adversaries benefit more from receiving features from other adversary devices when their shared knowledge is more limited. The function saturates rather quickly, highlighting the potency of the attack even for relatively sparsely connected graphs (i.e., $\rho \approx 1$). Therefore, if the adversaries decide to utilize Method 2 (Fig. \ref{fig:adver}) for their attack, a fully-connected graph is not necessary. Finally, we remark that when dealing with low connectivity values for MNIST and SVHN (i.e., $\rho = 0.1$), no datapoints are inferred via the consensus voting, leading to no datapoints being poisoned, i.e., no perturbation.

\section{Conclusion} 
In this paper, we introduced a novel methodology for conducting backdoor attacks in cross-device VFL environments. Our method considers a hybrid VAE and metric learning approach for label inference, and exploits the available graph topology among adversaries for cooperative trigger implantation. We theoretically analyzed VFL convergence behavior under backdooring, and showed that the server model would have a stationarity gap proportional to the level of adversarial gradient perturbation. Our numerical experiments showed that the proposed method surpasses existing baselines in label inference accuracy and attack performance across various datasets, while also exhibiting increased resilience to server-side defenses.

\vspace{-0.05in}
\appendix
\subsection{Proof of Theorem 1}
\label{proof:theorem}
First, we denote the mini-batch gradient of the loss function as   $$\nabla_{{\theta}^{(t)}} \mathcal{L} = [(\nabla_{\theta^{(t)}_{1}} \mathcal{L})^{\mathsf{T}}, (\nabla_{\theta^{(t)}_{2}}  \mathcal{L})^{\mathsf{T}}, \ldots, (\nabla_{\theta^{(t)}_{K}}  \mathcal{L})^{\mathsf{T}}]^{\mathsf{T}}.$$
Due to the adversarial attacks, the gradients used for the updates will be perturbed. We denote the perturbed gradient as $$\nabla^{a}_{{\theta}^{(t)}} \mathcal{L}=[(\nabla^{a}_{\theta^{(t)}_{1}} \mathcal{L})^{\mathsf{T}}, (\nabla^{a}_{\theta^{(t)}_{2}}  \mathcal{L})^{\mathsf{T}}, \ldots, (\nabla^{a}_{\theta^{(t)}_{K}}  \mathcal{L})^{\mathsf{T}}]^{\mathsf{T}},$$ which is further used for model update in VFL, i.e., $${\theta}_k^{(t+1)} =  {\theta}_k^{(t)} - \eta^{(t)} \nabla^{a}_{\theta^{(t)}_{k}}  \mathcal{L}.$$ Hence, the update of the whole model can be expressed as  
\begin{equation}
\begin{aligned}
{\theta}^{(t+1)} - {\theta}^{(t)} =& - \eta^{(t)}[(\nabla_{\theta^{(t)}_{1}} \mathcal{L})^{\mathsf{T}}, (\nabla_{\theta^{(t)}_{2}}  \mathcal{L})^{\mathsf{T}}, \ldots, (\nabla_{\theta^{(t)}_{K}}  \mathcal{L})^{\mathsf{T}}]^{\mathsf{T}} \\
&- \underbrace{[ \dots, (\nabla^a_{\theta^{(t)}_{k}}  \mathcal{L})^{\mathsf{T}} - (\nabla_{\theta^{(t)}_{k}}  \mathcal{L})^{\mathsf{T}}, \ldots]^{\mathsf{T}}}_{\Delta^{(t)}}.
\end{aligned}\nonumber
\end{equation}
From Assumption \ref{as:l-smooth} and the above iterative equation,  we have 
\begin{align}
\label{ineq:ftheta}
    & F({\theta}^{(t+1)}) \nonumber \\
    \leq & F({\theta}^{(t)}) \!+\! \langle \nabla F({\theta}^{(t)}), {\theta}^{(t+1)} \!-\! {\theta}^{(t)} \rangle \!+\! \frac{L}{2} \|{\theta}^{(t+1)} \!-\! {\theta}^{(t)}\|^2 \nonumber \\[-.1em]
    \leq & F({\theta}^{(t)}) - \langle \nabla F({\theta}^{(t)}), \eta^{(t)} \nabla_{{\theta}^{(t)}} \mathcal{L} + \eta^{(t)}\Delta^{(t)} \rangle \nonumber \\[-.1em]
    & + \frac{L}{2} \|\eta^{(t)} \nabla_{{\theta}^{(t)}} \mathcal{L} + \eta^{(t)}\Delta^{(t)}\|^2.
\end{align}
Next, from Assumption \ref{as:variance_grad}, we have $\mathbb{E}[\nabla_{\theta_{k}}  \mathcal{L} ] = \mathbb{E} [\nabla_{\theta_{k}}  F({\theta})]$. Therefore, taking expectation over \eqref{ineq:ftheta} leads us to
\begin{align}
    & \mathbb{E} [F({\theta}^{(t+1)})] \nonumber  \\
    \leq &  F({\theta}^{(t)}) - \eta^{(t)} \| \nabla F({\theta}^{(t)})\|^2 + (\eta^{(t)})^2 L \mathbb{E}\|\Delta^{(t)}\|^2\nonumber \\
    &   - \eta^{(t)} \mathbb{E}  \langle \nabla F({\theta}^{(t)}) ,\Delta^{(t)} \rangle  + (\eta^{(t)})^2 L \mathbb{E} \| \nabla_{{\theta}^{(t)}} \mathcal{L} \|^2  \nonumber \\
    \leq & F({\theta}^{(t)}) - \eta^{(t)} \| \nabla F({\theta}^{(t)})\|^2 + \frac{1}{2} \eta^{(t)} \| \nabla F({\theta}^{(t)})\|^2  \nonumber \\
    & + \frac{1}{2} \eta^{(t)} \mathbb{E} \|\Delta_t\|^2 + (\eta^{(t)})^2 L \mathbb{E} \| \nabla_{{\theta}^{(t)}} \mathcal{L} \|^2 \nonumber \\
    & + (\eta^{(t)})^2 L \mathbb{E} \|\Delta^{(t)}\|^2 \label{ineq:young} \\
    \leq & F({\theta}^{(t)}) - ( \eta^{(t)} -  \frac{1}{2} \eta^{(t)} - (\eta^{(t)})^2 L) \| \nabla F({\theta}^{(t)})\|^2   \nonumber \\
    &+ (\eta^{(t)})^2 L \mathbb{E} \| \nabla_{{\theta}^{(t)}} \mathcal{L} - \nabla F({\theta}^{(t)}) \|^2 \nonumber \\
    & + \frac{1}{2} (\eta^{(t)}) \mathbb{E} \|\Delta_t\|^2 + (\eta^{(t)})^2 L \mathbb{E} \|\Delta^{(t)}\|^2 \nonumber \\
    \leq & F({\theta}^{(t)})  - ( (\eta^{(t)}) -  \frac{1}{2} \eta^{(t)} - (\eta^{(t)})^2 L) \| \nabla F({\theta}^{(t)})\|^2  \nonumber \\
    &  + (\eta^{(t)})^2 KL \Gamma + \frac{1}{2} \eta^{(t)} K \delta (\rho) + (\eta^{(t)})^2 K L \delta (\rho), 
    \label{expansion_smooth}
\end{align}
where we utilize $\langle c_1, c_2 \rangle \leq \frac{\| c_1\|^2}{2} + \frac{\| c_2\|^2}{2}$ \cite{wen2022communication} to obtain \eqref{ineq:young}, and Assumptions 2 and 3 to obtain \eqref{expansion_smooth}. Now, letting $\eta^{(t)} \leq \frac{1}{4L}$, we have
$\eta^{(t)} \!-\!  \frac{1}{2} \eta^{(t)} \!-\! (\eta^{(t)})^2 L \!\geq \! \frac{\eta^{(t)}}{4}$. Hence, \eqref{expansion_smooth} can be rewritten as 
\begin{align}
  & \frac{\eta^{(t)}}{4}  \| \nabla F({\theta}^{(t)})\|^2 \\ \leq & F({\theta}^{(t)}) - \mathbb{E} [F({\theta}^{(t+1)})]  +  (\eta^{(t)})^2 K L \Gamma  \nonumber
  \\[-.1em]
  & + \frac{1}{2} \eta^{(t)} K \delta (\rho) + (\eta^{(t)})^2 K L \delta (\rho). \nonumber
\end{align}
Taking expectation of the above inequality over ${\theta}^{(t)}$, we have 
\begin{align}
  & \eta^{(t)} \mathbb{E} \| \nabla F({\theta}^{(t)})\|^2  \nonumber  \\
  \leq & 4 \mathbb{E} [F({\theta}^{(t)})] - 4 \mathbb{E} [F({\theta}^{(t+1)})]  + 4 (\eta^{(t)})^2 K L \Gamma  \nonumber \\
  &+ 4 (\eta^{(t)})^2 L K \delta (\rho) + 2 \eta^{(t)} K \delta (\rho) . \nonumber
\end{align}
Summing the above inequality from $t=0$ to $T-1$ and utilizing the fact that the loss function is non-negative, i.e., $\mathbb{E} [F({\theta}^T)] \geq 0$, we have
\begin{align}
 & \sum_{t=0}^{T-1} \eta^{(t)} \mathbb{E} \| \nabla F({\theta}^{(t)})\|^2  
  \leq 4 F({\theta}^0)  + 4  (\sum_{t=0}^{T-1}(\eta^{(t)})^2) K L \Gamma \nonumber \\
  & + 4 (\sum_{t=0}^{T-1}(\eta^{(t)})^2) L K \delta (\rho) + 2 (\sum_{t=0}^{T-1}\eta^{(t)}) K \delta (\rho). 
\end{align}
Dividing the both sides by $\sum_{t=0}^{T-1} \eta^{(t)}$ leads us to
\begin{align}
  & \sum_{t=0}^{T-1} \frac{\eta^{(t)}}{\sum_{t=0}^{T-1} \eta^{(t)}} \mathbb{E} \| \nabla F({\theta}^{(t)})\|^2  \nonumber \\
  \leq & 4 \frac{F({\theta}^0)}{\sum_{t=0}^{T-1} \eta^{(t)}}  + 4  (\sum_{t=0}^{T-1}(\eta^{(t)})) K L \Gamma \nonumber \\
  & + 4 (\sum_{t=0}^{T-1}(\eta^{(t)})) L K \delta (\rho) + 2 K \delta (\rho). 
\end{align}
Furthermore, as $\min_{t=0,\ldots,T-1} z_t \leq \frac{\omega_t}{\sum_{t=0}^{T-1} \omega_t} z_t$, we thus complete the proof of Theorem 1.

\balance
\bibliographystyle{IEEEtran}
\bibliography{references.bib}

\end{document}